\documentclass[11pt]{article}

\usepackage[final]{acl}

\usepackage{times}
\usepackage{latexsym}
\usepackage{enumitem}
\usepackage[T1]{fontenc}

\usepackage[utf8]{inputenc}
\usepackage{amsfonts}
\usepackage{graphicx}
\usepackage{microtype}

\usepackage{inconsolata}

\usepackage{makecell}
\usepackage{graphicx}
\usepackage{booktabs}
\usepackage{multirow}
\usepackage{amsmath} 
\title{Surgical Alignment in Knowledge Graph Training for Clinical Diagnosis with Large Language Models}

\author{
 \textbf{Saksham Khatwani\textsuperscript{1}},
 \textbf{He Cheng\textsuperscript{1}},
 \textbf{Majid Afshar\textsuperscript{2}},
 \textbf{Dmitriy Dligach\textsuperscript{3}},
 \textbf{Yanjun Gao\textsuperscript{1}}
\\
 \textsuperscript{1}University of Colorado Anschutz Medical Campus,
 \textsuperscript{2}University of Wisconsin Madison,\
\\
 \textsuperscript{3}Loyola University
\\
 \small{
   \textbf{Correspondence:} \href{mailto:yanjun.gao@cuanschutz.edu}{yanjun.gao@cuanschutz.edu}
 }
}

\begin{document}
\maketitle
\begin{abstract}
Biomedical knowledge graphs (KGs) offer structured medical knowledge that can ground large language model (LLM) reasoning in clinical diagnosis application, yet how KG signal should be integrated into LLMs remains an open question. We present a systematic study spanning five KG task formulations, three training paradigms, two KGs, and three base LLMs. At the task level, all paradigms improve over the non-finetuned baseline, but methods with comparable in-domain accuracy show substantially different knowledge transfer behavior. We introduce \textbf{Gradient Intervention Density (GID)} and \textbf{Gradient Distortion (GD)} to 
measure how broadly an optimizer modifies the pretrained model. GID and GD together reveal a clear divide: KG-judgment training under KL regularization produces sparse, localized updates (a regime we term as \textit{surgical alignment}), while task-specific SFT produces dense ones. A controlled ablation shows that the objective and KL contribute to sparsity independently, and the paradigms that produce sparse updates also improve reasoning quality, even when their in-domain accuracy is lower than task-specific SFT. Assessing KG-LLM integration thus requires complementing accuracy with optimization-geometry diagnostics. Our implementation can be found at \url{https://github.com/LARK-NLP-Lab/Surgical-Alignment}.
\end{abstract}

\section{Introduction}
Large language models (LLMs) have demonstrated remarkable potential in clinical diagnostics~\cite{goh2024large,wang2025medical}, yet their outputs remain prone to factual error rather than grounded medical logic~\cite{kim2025medical,asgari2025framework}. Biomedical Knowledge Graphs (KGs), which encode clinical concepts and their relations in a curated, structured form, offer a natural scaffolding necessary for trustworthy reasoning for grounding LLM reasoning in medical knowledge. However, how to incorporate KGs into LLMs for clinical diagnosis remains an open question. 

Existing approaches differ in where the KG enters the pipeline. Inference-time methods use KG as retrieval context (e.g. Retrieval-Augmented Generation with prompting-based method~\cite{gao2025leveraging, zuo2025kg4diagnosis}, or with external memories for agentic systems~\cite{10.24963/ijcai.2025/2,xu-etal-2025-memory}). Training-time methods supervise LLM directly on KG-derived signals, either through supervised fine-tuning of the graph structure~\cite{tian-etal-2024-kg,chen2025knowledge,chen2025reasongrm}, or through preference-based reinforcement learning (RL) such as Group Relative Policy Optimization (GRPO)~\cite{shao2024deepseekmathpushinglimitsmathematical}, reasoning as a reward model~\cite{yan2025rlkgf,han2025reasoning}, and implicit rewards derived from KG~\cite{kansal2026knowledge}. Within training-time methods, approaches further differ in task formulation: models may be trained to judge candidate reasoning paths or to generate paths from partial input~\cite{kansal2026knowledge}.
As existing work is largely optimized and evaluated on narrow benchmarks and task-specific targets, the field lacks a unified understanding of which task formulations and objectives produce KG-grounded reasoning that generalizes beyond the supervision task, particularly in clinical diagnosis where KG reasoning may improve faithfulness.  

We focus on the training-time question through a systematic study spanning five KG task formulations, consisting of three judgment-based tasks that assess the validity of reasoning paths and two generative tasks that complete partial reasoning chains. Moreover, we study three training paradigms, two KGs (UMLS~\cite{Bodenreider2004TheUM} and PrimeKG~\cite{Chandak2023-qj}), and three base LLMs (Qwen2.5-7B, Qwen3-8B, Gemma-7B)~\cite{qwen2025qwen25technicalreport,yang2025qwen3technicalreport, gemmateam2024gemmaopenmodelsbased}. We investigate two complementary questions: (i) how each combination of task formulation and  training paradigm shapes transfer to other KG tasks and downstream clinical benchmarks, and (ii) how each training objective reshapes the pretrained model itself. For the first, we evaluate transfer using ROUGE-L~\cite{lin2004rouge}, CUI-F~\cite{afshar2024role, gao-etal-2022-summarizing}, and QA accuracy across diagnosis prediction (ProbSum, DDXPlus) \cite{gao-etal-2022-summarizing, tchango2022ddxplusnewdatasetautomatic} and medical QA (MedQA, MedMCQA)~\cite{jin2020diseasedoespatienthave,pal2022medmcqalargescalemultisubject}. All paradigms improve over the non-finetuned baseline, but methods with comparable in-domain accuracy exhibit substantially different transfer behavior, and their relative ordering reverses across benchmarks. 

For the second, we move beneath task performance, since accuracy  alone \textit{does not} reveal how a given objective actually reshapes the model. We introduce \textbf{Gradient Intervention Density (GID)}, a layer-wise diagnostic that quantifies the sparsity of an optimizer's update footprint relative to the pretrained baseline. Additionally, we define \textbf{Gradient Distortion (GD)}, which quantifies the deviation of model parameters from their pretrained baseline. Judgment-based KG objectives under Kullback-Leibler (KL) regularization induce sparse, localized updates, a regime we term \textit{surgical alignment}, whereas task-specific SFT produces dense ones. A controlled ablation study shows that the KG objective alone produces sparser updates than task-specific SFT (even without 
KL), and also improves clinical reasoning quality evaluated by PDSQI-9~\citep{croxford2025developmentvalidationproviderdocumentation}.  

Our proposal is \textit{not} a new RL architecture, but a novel analytical framework and insights for KG-LLM integration: the impact of different training paradigm on KGs, and what changes go behind it. Optimization geometry, alongside task accuracy, becomes a first-class evaluation dimension. While we evaluate this framework in clinical use-cases, it is domain agnostic. Our contributions include:
\begin{itemize}[leftmargin=*,noitemsep,topsep=0pt]
    \item \textbf{Empirical landscape of KG-LLM integration} We present a comprehensive study of various KG task formulations (path-judging vs. path generation), training paradigms (supervised fine-tuning vs. reinforcement learning), KG and LLM choice, examining how each combination shapes knowledge transfer across KG tasks and clinical diagnostic benchmarks. 
    \item \textbf{A lens of optimization-geometry} We provide the first application of  gradient-level optimization-geometry analysis to KG-LLM integration. We introduce \textbf{Gradient Intervention Density (GID)} and \textbf{Gradient Distortion (GD)}, layer-wise metrics that quantifies the sparsity and deviation of an optimizer's update footprint relative to the pretrained baseline. 
    \item \textbf{Conceptualizing surgical alignment} Combining GID and GD analysis with a controlled ablation that varies training objective and KL strength independently, we establish \textit{surgical alignment}: a regime where judgment-based KG training produces sparse, localized parameter updates that co-occur with improvements on higher-level PDSQI-9 reasoning dimensions (organization, synthesis), even when in-domain accuracy is lower than task-specific SFT.  
\end{itemize}

\begin{figure*}[t]
    \centering
    \includegraphics[width=\linewidth]{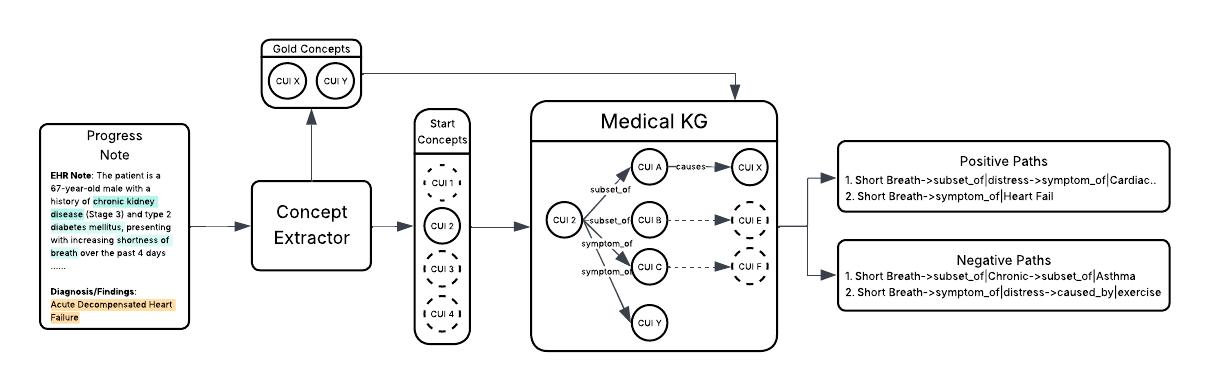}
    \vspace{-.39in}
    \caption{\small KG path extraction from patient progress notes. The concept extractor (\textsc{QuickUMLS}~\cite{Soldaini2016QuickUMLSAF} for UMLS KG and \textsc{Simstring-fast}~\cite{okazaki-tsujii-2010-simple} for PrimeKG) identifies the starting concepts based on entities in progress notes, and gold concepts based on diagnoses labels. For example in this figure, we can see that starting from concept \textsc{CUI-2}, the paths which end with \textsc{CUI-X} and \textsc{CUI-Y} are positive samples and other paths are negative.}
    \label{fig:path_extraction_workflow}
    \vspace{-.1in}
\end{figure*}

\section{Related Work}




\textit{KG as Context.}
KG can serve as dynamic context during inference through retrieval, prompting, and memory. 
Beyond the work discussed in \S 1, \citet{jia2024medikal} generate and rank candidate diagnoses using KG knowledge (\textit{medIKAL}), while \citet{zhao2025medrag} combine RAG with a structured diagnostic KG (\textit{MedRAG}) to refine EHR-based predictions.
\citet{zhang2024knowgpt} introduce \textit{KnowGPT}, which extracts relevant subgraphs to construct context-aware prompts. \citet{lee2025chain} propose \textit{Chain of Knowledge Graph (CoKG)}, where models construct and validate a weighted KG before summarization. Other works include evidence-grounded retrieval and reasoning frameworks such as \citet{wu2024medicalgraphragsafe} and \citet{Jiang2024ReasoningEnhancedHP}.
KGs have also been used as structured memory for long-term reasoning: \citet{jiang2026magma} organize memory into explicit graphs for precise retrieval, and \citet{10.24963/ijcai.2025/2} build evolving experience-based KGs for planning. These methods treat KGs as external context rather than explicit training signals.

\textit{KG as Training Supervision.}
Recent work use KG-based supervision signals for training. KG-Adapter~\cite{tian-etal-2024-kg} injects KG structure via parameter-efficient adapters and requires architecture modification.~\citeauthor{kansal2026knowledge} (\citeyear{kansal2026knowledge}) use KG paths as implicit reward models for RL, enabling compositional reasoning and generalization to longer reasoning chains. \citeauthor{chen2025knowledge} (\citeyear{chen2025knowledge}) demonstrate that KG-enhanced fine-tuning improves knowledge manipulation capabilities in low-data scenarios. These methods prove that KG can serve as effective signals for training.

\textit{Reward Models as Reasoning.}
Recent work reframes reward models as implicit reasoners rather than mere scorers. \textit{RM-R1}~\cite{rmr1} treats reward modeling as a reasoning process with explicit rationales, while \textit{ReasonGRM}~\cite{chen2025reasongrm} uses large reasoning models to generate detailed reward explanations. \citeauthor{guo2025reward} (\citeyear{guo2025reward}) propose reasoning reward models that perform explicit inference without ground-truth traces, and \citet{prm_think2024} introduce step-by-step reasoning verifiers.

\section{Data Overview}
\textbf{KG selection.} We use two KGs, UMLS~\citep{Bodenreider2004TheUM} 
and PrimeKG~\citep{Chandak2023-qj}. For UMLS, we adopt the  diagnostic-reasoning version from \citet{gao2025leveraging}, retaining 107 SNOMED CT-based semantic relations relevant to diagnosis. For PrimeKG, we focus on Disease, Phenotype, and Drug 
node types.


\textbf{Training data.} We use two clinical datasets: \textbf{ProbSum} \cite{gao-etal-2023-overview} and \textbf{DDXPlus} \cite{tchango2022ddxplusnewdatasetautomatic}. The ProbSum dataset was introduced as part of the BioNLP 2023 shared task on patient summarization \cite{gao-etal-2023-overview}. It contains 1,005 physician-annotated progress notes of ICU patients from MIMIC-III \cite{Johnson2016MIMICIIIAF}. DDX-Plus is a large scale synthetic dataset which includes differential diagnosis, along with ground truth pathology, and antecedents for each patient. The dataset covers 49 pathologies, 110 symptoms and 113 antecedents. 

\textbf{Downstream evaluation.} We use the \textbf{MedQA}~\cite{jin2020diseasedoespatienthave} and \textbf{MedMCQA}~\cite{pal2022medmcqalargescalemultisubject} datasets. MedQA contains multiple-choice medical questions with four answer options and one correct answer. It spans various clinical scenarios, including treatment, management, and diagnosis. For our study, we focused on diagnosis-related questions, yielding 1,796 training and 251 test samples. MedMCQA dataset is also a multiple-choice medical QA dataset derived from Indian medical entrance exams spanning various medical scenarios. We limit our study to the 2,560 diagnosis related questions.

\begin{figure*}[t!]
    \centering
    \hspace*{-1.3cm}%
     \includegraphics[width=0.98\textwidth]{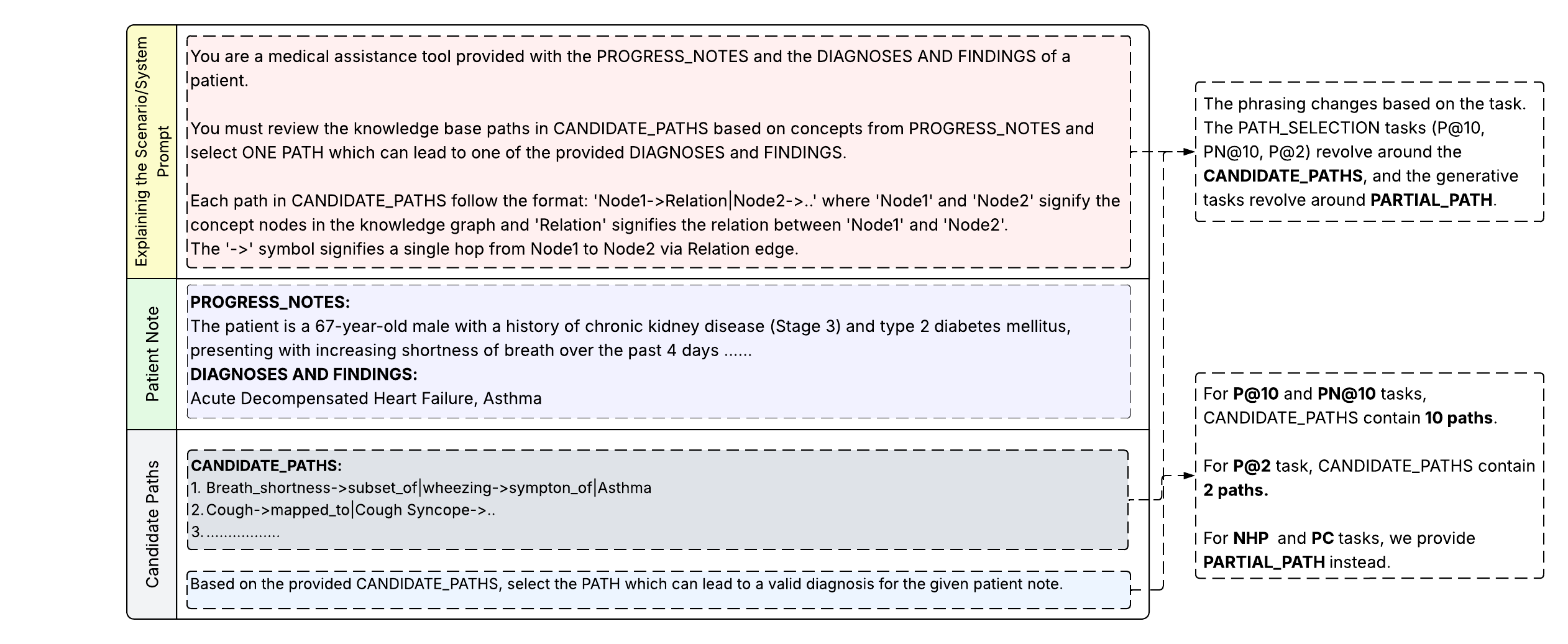}
     \vspace{-.1in}
    \caption{\small Training Prompt structure for the different task formulation. The basic format includes the system initialization explaining the scenario, patient progress notes, patient diagnoses, either \textbf{CANDIDATE\_PATH} or \textbf{PARTIAL\_PATH} based on the task, and the final line explaining the task. For evaluation, we follow the same structure of the prompt, but remove the patient diagnoses part. CUI denotes Concept Unique Identifier, representing the KG node which our model's output maps to.  }
    \label{fig:prompt_structure}
    \vspace{-.2in}
\end{figure*}

\section{Preprocessing} 
\label{Methods}

\noindent \textbf{KG Path Extraction}
For each patient note, we map entities in the note to KG nodes (starting concepts) and the gold diagnosis to KG nodes (target concepts), then perform breadth-first search up to two hops from each starting concept. Paths reaching a target concept are labeled \textbf{positive}; others \textbf{negative} (as presented in Figure~\ref{fig:path_extraction_workflow}). For UMLS, we use QuickUMLS~\cite{Soldaini2016QuickUMLSAF} and retain only diagnostically relevant semantic types (e.g., diseases, findings, and symptoms). In case of PrimeKG, we map entities to KG nodes through the \textsc{SIMSTRING-FAST} library~\cite{okazaki-tsujii-2010-simple} and focus on the Disease, Phenotype, and Drug node types. Concept Unique Identifiers (CUIs) anchor cross-KG alignment but are not provided as model input; they are used only for concept-level evaluation. 
\begin{table}[h]
\centering
\small
\resizebox{0.48\textwidth}{!}{
\begin{tabular}{@{}lll}
\toprule \toprule
\textbf{Task Type} & \textbf{Task} & \textbf{Description} \\ 
\midrule
\multirow{3}{*}[-1em]{Path Selection} 
    & \textsc{P@10} & \makecell[l]{Given 10 candidate KG paths, \\ identify one valid path.} \\ \cmidrule{2-3}
    & \textsc{P@2}  & \makecell[l]{Given 2 candidate KG paths, \\ identify one valid path.} \\ \cmidrule{2-3}
    & \textsc{PN@10} & \makecell[l]{Given 10 candidate KG paths, \\ identify multiple valid paths.} \\
\midrule
\multirow{1}{*}{Next-Hop} 
    & \textsc{NHP} & \makecell[l]{Given a partial KG path, \\ predict the next hop.} \\ \midrule
\multirow{1}{*}{Path Completion} 
    & \textsc{PC} & \makecell[l]{Given a partial KG path, \\ predict the remaining path.} \\
\bottomrule \bottomrule
\end{tabular}
}
\vspace{-.13in}
\caption{\small Task formulations for training LLMs to reason over medical knowledge graphs. The three task types (Path Selection, NHP, and PC) probe complementary reasoning abilities. The Path Selection task requires model to judge the validity of the reasoning paths, whereas NHP and PC are generative reasoning tasks which require the models to complete the partial reasoning chains.}
\vspace{-.13in}
\label{tab:task_formulation}
\end{table}



\noindent \textbf{KG Training Task Formulation} We define three task formulations over patient-specific KG paths.  \textsc{Path Selection} judges valid reasoning paths from candidate sets, with three variants: \textsc{P@10}, \textsc{P@2}, and \textsc{PN@10}. \textsc{Next-Hop Prediction (NHP)} and  \textsc{Path Completion (PC)} are generative: given a partial path, the model predicts the next hop or completes the remaining path, respectively. See Table~\ref{tab:task_formulation} for task details and Fig~\ref{fig:prompt_structure} for prompt structure.


\section{KG Training Paradigm} 
We conduct our experiments using Qwen2.5-7B-Instruct (Qwn7B)\cite{qwen2025qwen25technicalreport}, Qwen3-8B (Qwn8B)\cite{yang2025qwen3technicalreport}, and Gemma-7B-IT (Gem7B)\cite{gemmateam2024gemmaopenmodelsbased}. 

\subsection{KG-based SFT} 
\label{sec:kg-based-sft}

We perform supervised fine-tuning (SFT) under each individual task formulation listed in Table \ref{tab:task_formulation}. SFT serves as an initial alignment stage that exposes the model to KG structure; we treat any resulting memorization of graph patterns as an expected property of this stage. In addition, a multi-task SFT is applied to provide more uniform exposure to diverse KG structures and task formulations, reducing bias toward any single reasoning pattern. Concretely, we pool training data across the five task formulations in Table~\ref{tab:task_formulation}, subsampled to keep per-task training size consistent. 

 To preserve the base model's output distribution, we add a 
Kullback-Leibler (KL) regularization term to the cross-entropy 
loss:

\vspace{-.2in}
\begin{equation}
\small
\label{eq:grpo}
\begin{split}
\mathcal{L}_{\text{total}}(\theta)
=
\sum_{t=1}^{T}
\Bigg(
- \log p_\theta\!\left(y_t \mid y_{<t}, x\right)
+ \\ \lambda \cdot \,
\mathrm{KL}
\Big(
p_\theta(y_t \mid y_{<t}, x)
\;\|\;
p_{\text{NFT}}(y_t \mid y_{<t}, x)
\Big)
\Bigg), \\
\end{split}
\end{equation}

where $\lambda \in [0.01, 1]$ controls KL strength and 
$p_{\text{NFT}}$ denotes the non-finetuned baseline distribution. 


\subsection{KG Reward Model as Reasoning Training }

Beyond direct token-level fitting via SFT, reward-based training frames KG learning as preference optimization over candidate outputs, which prior work suggests can yield more structured reasoning behavior~\cite{rmr1,shao2024deepseekmathpushinglimitsmathematical}. We instantiate this in three paradigms: GRPO, RM-R1, and Comp-GRPO.

\paragraph{GRPO.} 
Each training instance contains a group of candidate paths (e.g., 10 paths for P@10, or PN@10); GRPO~\cite{shao2024deepseekmathpushinglimitsmathematical} optimizes relative preferences by assigning higher likelihood to valid diagnostic paths across the group. We apply GRPO to models that have already undergone KG-based SFT for each task formulation (as in \S \ref{sec:kg-based-sft}). Importantly, because GRPO optimizes relative preferences within candidate groups under KL regularization, it encourages localized, comparison-driven updates rather than dense token-level fitting.

The reward function differs by task. For Path Selection, the reward is positive only when the model output contains all valid paths and none of the invalid ones:
\vspace{-.09in}
\begin{equation}
\small
\mathcal{R}(y) =
\begin{cases} 
a, & \forall \hat{y} \in \text{valid\_paths},\ \hat{y} \in y \\
   & \&\ \forall \hat{y} \in \text{wrong\_paths},\ \hat{y} \notin y \\
b, & \text{otherwise}
\end{cases}
\label{eq:reward-pathsel}
\end{equation}

For NHP, where multiple text outputs may refer to the same KG node, we map predictions to KG nodes via the concept extractors and reward CUI-level matching:
\vspace{-.09in}
\begin{equation}
\small
\mathcal{R}(y) =
\begin{cases} 
a, & \text{if CUI}(y) = \text{CUI(true label)} \\
b, & \text{otherwise}
\end{cases}
\label{eq:reward-nhp}
\end{equation}


For PC task, we reward positively if the model's path completion is equal to the label completion:
\vspace{-.08in}
\begin{equation}
\small
\mathcal{R}(y) =
\begin{cases} 
a, & \text{if } y = \text{true label} \\
b, & \text{otherwise}
\end{cases}
\label{eq:pc-reward}
\end{equation}

In our experiments, we set $a=1$ and $b=0$. For the multi-task GRPO training, our reward function first checks the task type and then assigns the reward accordingly.


\paragraph{RM-R1} We adopt the RM-R1 framework~\cite{rmr1} 
for training reasoning reward models (\textsc{ReasRM}) in two  stages: (1) reasoning distillation via SFT on Chain-of-Thought traces, and (2) GRPO over the distilled reward model. For SFT data, we sample 500 instances per task formulation and prompt HIPAA-compliant Azure GPT-o3-mini with patient notes, gold diagnoses, and the task answer to elicit detailed justifications. The RL stage reuses the GRPO reward (Eq. \ref{eq:pc-reward}) with $a=1$, 
$b=-1$, per the original paper. 

\noindent \textbf{Comp-GRPO} We utilize the SFT+RL framework introduced by~\cite{kansal2026knowledge} to train compositional KG-grounded reasoning models. For the SFT stage, we utilize our KG-based SFT checkpoints. The RL stage of this framework assigns a discrete reward ($R_{bin}$) and a path alignment reward ($R_{path}$) based on the token overlap between the model's reasoning trace and the ground truth path. For $R_{bin}$, we use the previously defined reward function (Eq. \ref{eq:pc-reward}) with $a=0.1$ and $b=-1$ as defined in the framework. For $R_{path}$, in accordance to the framework, we first tokenize and normalize the model's final answer (y) to extract textual tokens $T(y)$. We also tokenize the ground truth label for the task and extract ground truth tokens $T(gold)$. The coverage between the two sets of textual tokens is defined as:

\begin{equation}
\small
\text{coverage}
= \frac{|T(y) \cap T(gold)|}{|T(gold)|}
\end{equation}

We define a minimum coverage constraint of at least two distinct tokens and apply a repetition penalty to mitigate reward hacking. $R_{path}$ is defined as:

\begin{equation}
\small
\begin{aligned}
R_{\text{path}} (y, gold) =
\min \Big(
\gamma_1 \cdot \mathrm{coverage}
+ \\
\gamma_2 \cdot \mathbb{I}\big(|T(y) \cap T(gold)| \geq 2\big),
\; R_{\max}
\Big)
\end{aligned}
\end{equation}

Where $\gamma_1$= 1.2, $\gamma_2$ = 0.3, $R_{\max}$ = 1.5 as defined by \cite{kansal2026knowledge}. This reward is scaled by repetition penalty factor.

We conduct this training in two settings: (1) Applying both $R_{bin}$ and $R_{path}$, and (2) Applying only $R_{path}$.

\subsection{Evaluation Setup}
\paragraph{Metrics.}  We use ROUGE-L~\cite{lin2004rouge} for lexical overlap, and 
CUI-F~\cite{afshar2024role, gao-etal-2022-summarizing} for 
concept-level overlap by mapping generated text back to KG CUIs 
via QuickUMLS (UMLS) and SIMSTRING (PrimeKG); precision/recall 
formulation is in Appendix~\ref{appendix: evaluation_setup_additional}. ROUGE-L and CUI-F 
apply to KG-path tasks and to diagnosis prediction (ProbSum, 
DDXPlus); MedQA and MedMCQA use exact matching. We additionally 
evaluate diagnostic reasoning quality on ProbSum using 
PDSQI-9~\cite{croxford2025developmentvalidationproviderdocumentation}, 
a nine-criteria clinical rubric scored by an Azure-hosted GPT-5-mini 
judge previously shown to correlate highly with human ratings.


\paragraph{Baselines.}
We compare against three baselines. (i) \textbf{Task-specific SFT}: 
LLMs fine-tuned directly on ProbSum and DDXPlus and evaluated 
across all downstream tasks. We do not fine-tune on QA benchmarks 
to avoid leaderboard contamination with pretraining data. 
(ii) \textbf{RAG (UMLS)} and \textbf{RAG (PrimeKG)}: at inference 
time, we extract disease/symptom concepts from patient notes (or 
map QA answer choices to KG nodes) and retrieve 10 paths via BFS 
up to 2 hops, prepended to the NFT model's prompt. 
(iii) \textbf{NFT}: the non-finetuned base model. We provide further details regarding how the models were trained, hardware, quantization, and Low-Rank Adaptation (LoRA)~\cite{hu2021loralowrankadaptationlarge} in appendix~\ref{training_details}.

\begin{figure}[t!] 

   \centering
   \includegraphics[width=\columnwidth]{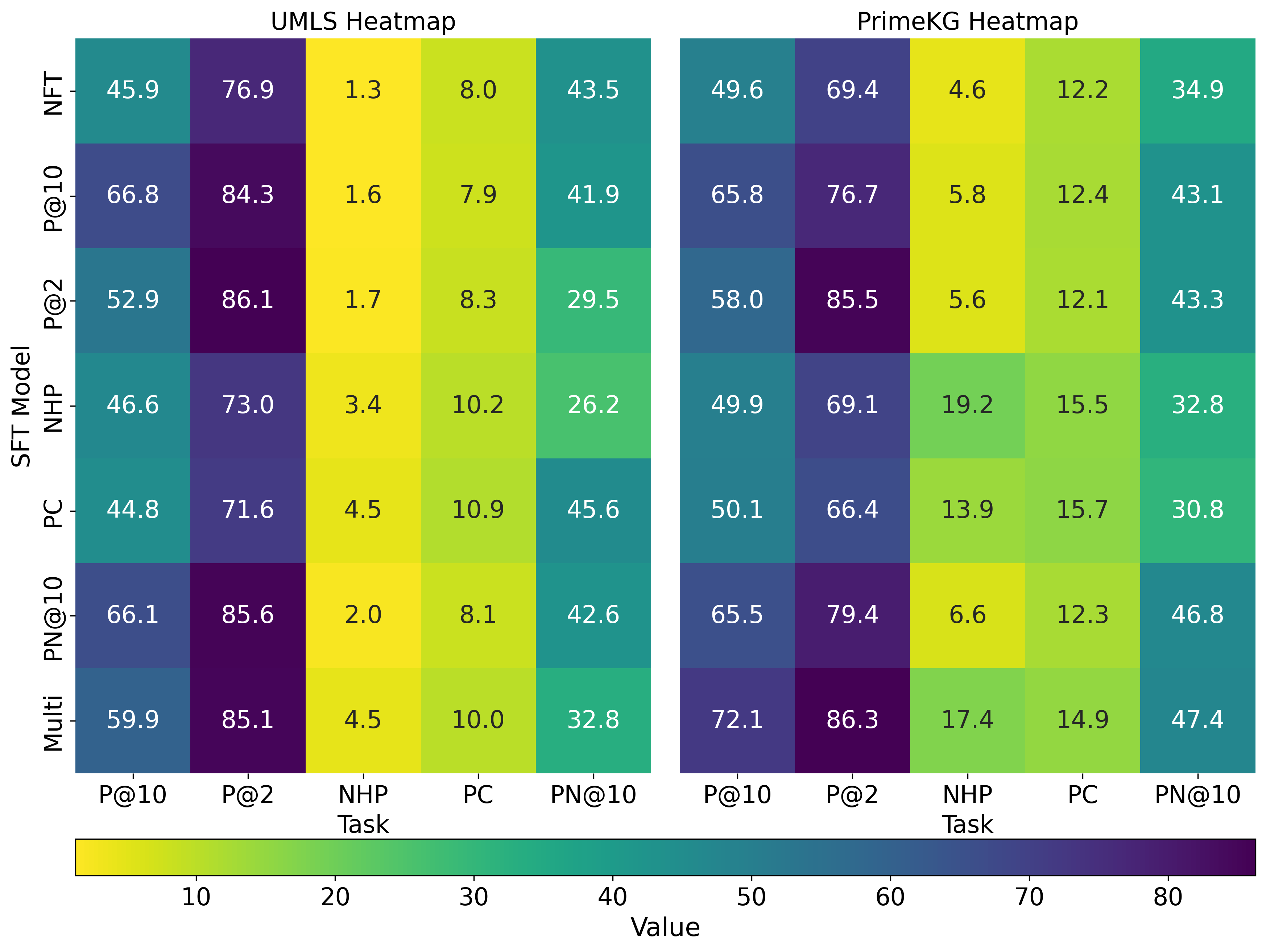}
   \vspace{-.32in}
    \caption{\small Cross-task generalization among the path-judging formulations across UMLS and PrimeKG. Results are reported in ROUGE-L scores. Due to space constraints, we present results from Qwen2.5-7B-Instruct, but observe consistent patterns across the other LLMs.  }
    \vspace{-.2in}
    \label{fig: task_heatmap}
\end{figure}
    
 \begin{table*}[htbp]
     \scriptsize 
    \centering
    
    \resizebox{\textwidth}{!}{
    \begin{tabular}{lcccccccccccc}
    
    \toprule \toprule
     &  \multicolumn{6}{c}{UMLS} & \multicolumn{6}{c}{PrimeKG} \\
    \cmidrule(lr){2-7} \cmidrule(lr){8-13} 
        \multirow{2}{*}{Training Paradigm} & \multicolumn{2}{c}{Qwn7B} & \multicolumn{2}{c}{Qwn8B} & \multicolumn{2}{c}{Gem7B} & \multicolumn{2}{c}{Qwn7B} & \multicolumn{2}{c}{Qwn8B} & \multicolumn{2}{c}{Gem7B} \\ \cmidrule(lr){2-7} \cmidrule(lr){8-13} 
                                            & RL & CF & RL & CF & RL & CF & RL & CF & RL & CF & RL & CF \\
        \midrule
        {NFT}   &46.14 & 51.29& 36.42& 39.26& 24.70& 28.84&49.55 &37.23&32.70 &25.03 &45.99 &40.37 \\ 
         
        {SFT}& 59.89&60.92 & 44.71 & 45.90 &\textbf{68.03} &\textbf{69.68} & 63.34& 64.73&47.16 &41.90 &\textbf{71.95}&\textbf{65.23}\\
         
        GRPO   & 57.74& 59.24& 56.82& 58.29& 64.37& 66.34& 68.09& 60.95& 49.66& 40.81& 71.73& 65.75 \\ 
         
         Comp-GRPO   & 59.30 & 61.60&53.40& 54.62 & 61.23& 63.08& 68.62& 61.13& 51.93&43.95 &71.89 &65.75  \\ 
         
         Comp-GRPO (only $R_{path}$)  & \textbf{60.32}& \textbf{62.13}& 55.12 & 55.76 & 59.70 & 61.94 &\textbf{69.31} &6\textbf{0.95} & 51.80& 42.69& 71.60& 65.43 \\ 
         
        {RM-R1}  & 58.04& 58.79& \textbf{63.91}& \textbf{62.75}& 61.33& 62.01& 61.94& 60.78&\textbf{60.58}&\textbf{59.13} &71.29 & 71.70\\ 

        \bottomrule \bottomrule
        
    \end{tabular}
    }
    \vspace{-.1in}
    \caption{\small Rouge-L and CUI-F scores comparing various graph training paradigm across Qwen2.5-7B-Instruct, Qwen3-8B, and Gemma-7B-IT. We compare the multi-task checkpoints for the KG-based SFT, GRPO, and RM-R1 training paradigms against the NFT baseline. GRPO is performed to models which have already undergone SFT on KG paths. Comparison between different task formulations is highlighted in figure \ref{fig: task_heatmap}. The best performing checkpoints for each KG are marked with bold text. }
    \label{tab:training_paradigm}
\end{table*}

\begin{table*}[t]
    \scriptsize
    \centering
    \resizebox{\textwidth}{!}{
\begin{tabular}{llcccccccccccc}
    \toprule \toprule
        & \multirow{3}{*}{Training} & \multicolumn{6}{c}{ProbSum} & \multicolumn{6}{c}{DDXPlus} \\
        \cmidrule(lr){3-8} \cmidrule(lr){9-14}
        & & \multicolumn{2}{c}{Qwn7B} & \multicolumn{2}{c}{Qwn8B} & \multicolumn{2}{c}{Gem7B} & \multicolumn{2}{c}{Qwn7B} & \multicolumn{2}{c}{Qwn8B} & \multicolumn{2}{c}{Gem7B} \\ \cmidrule(lr){3-8} \cmidrule(lr){9-14} 
        &  & RL & CF & RL & CF & RL & CF & RL & CF & RL & CF & RL & CF \\
        \midrule
        \multirow{3}{*}{Baseline}
         & NFT & 22.14 & 26.05 & 04.15 & 16.36 & 20.20 & 25.33 & 10.90 & 09.50 & 01.58 & 05.72 & 06.38 & 05.52 \\
         & SFT & 26.09 & 27.53 & 04.72 & 17.87 & 17.48 & 30.42 & 22.98 & 46.20 & 13.62 & 38.95 & 28.60 & 46.07 \\
         & Cross SFT & 03.89 & 08.23 & 03.79 & 10.43 & 02.91 & 03.76 & 12.52 & 08.98 & 01.62 & 05.95 & 04.60 & 04.72 \\
         & RAG - UMLS & 14.09 & 06.09 & 09.08 & 03.05 & 08.49 & 14.65 & 05.45 & 02.66 & 04.00 & 01.59 & 03.37 & 07.17 \\
         & RAG - PrimeKG & 12.23 & 04.89 & 06.50 & 01.00 & 10.90 & 15.59 & 02.89 & 01.06 & 02.96 & 00.81 & 03.10 & 06.88 \\
        \midrule
        \multirow{5}{*}{\makecell[l]{UMLS}}
        & SFT & 22.79 & 26.11 & 04.16 & 16.33 & \textbf{19.33} & \textbf{26.98} & 11.75 & 08.64 & \textbf{01.55} & \textbf{05.71} & 06.93 & 05.50 \\
        & GRPO & 22.96 & 26.28 & 04.19 & 16.57 & 19.14 & 26.35 & 11.75 & 08.64 & 01.55 & 05.58 & 06.98 & 05.52 \\
        & Comp-GRPO & 20.82 & 19.40 & 04.78 & 27.26 & 14.01 & 26.82 & 10.47 & 10.44 & 01.26 & 06.90 & 05.51 & 08.28 \\
        & Comp-GRPO (only $R_{path}$) & \textbf{23.92} & \textbf{22.23} & \textbf{19.01} & \textbf{04.12} & 14.25 & 28.97 & 10.54 & 10.52 & 01.26 & 06.95 & 05.46 & 08.10 \\
        & RM-R1 & 21.21 & 24.13 & 04.33 & 17.25 & 19.18 & 23.93 & \textbf{12.27} & \textbf{08.99} & 01.40 & 04.98 & \textbf{09.70} & \textbf{08.47} \\
        \midrule
        \multirow{5}{*}{\makecell[l]{PrimeKG}}
         & SFT & 22.07 & 26.52 & 04.18 & 16.55 & 16.79 & 24.73 & 11.08 & 09.52 & 01.55 & 05.59 & 07.06 & 06.15 \\
        & GRPO & \textbf{22.09} & \textbf{26.31} & 04.28 & 16.91 & 16.61 & 25.11 & 11.14 & 09.49 & \textbf{01.57} & \textbf{05.69} & 07.01 & 06.11 \\
        & Comp-GRPO & 19.95 & 23.77 & 04.90 & 26.95 & 12.50 & 26.55 & 10.05 & 09.60 & 01.24 & 07.03 & 05.34 & 08.39 \\
        & Comp-GRPO (only $R_{path}$) & 19.52 & 22.57 & \textbf{04.90} & \textbf{27.42} & 12.58 & 25.71 & 10.03 & 09.66 & 01.25 & 07.01 & 05.35 & 08.52 \\
        & RM-R1 & 20.52 & 21.96 & 04.58 & 17.29 & \textbf{19.74} & \textbf{21.44} & \textbf{12.44} & \textbf{11.02} & 01.48 & 05.34 & \textbf{10.88} & \textbf{09.18} \\
        \bottomrule \bottomrule
\end{tabular}}
\vspace{-.1in}
\caption{\small Rouge-L (RL) and CUI-F (CF) scores for diagnosis prediction on ProbSum and DDXPlus test set. Multi-task checkpoints are evaluated alongside baseline methods. Best performing KG-trained method for each KG is marked as bold.}
\label{tab: probsum_ddxplus_rouge_cui}
    \vspace{-.1in}
\end{table*}

\begin{table}[t]
    \small
    \setlength{\tabcolsep}{2pt}
    \begin{tabular}{llccc}
    \toprule \toprule
        Task & Training & Qwn7B & Qwn8B & GemB \\ \midrule 
        \multirow{6}{*}{MedMCQA} &NFT               & 57.14& 36.73& 36.73 \\
        &RAG - UMLS               & 54.42& 36.73& 36.73 \\
        &RAG - PrimeKG               & 54.42& 36.73& 36.73 \\

        &Probsum SFT                    &  \textbf{63.27}& 36.73 & 32.65\\

        &DDXPlus SFT                    &  63.27& 32.65& 30.61\\

        &\makecell[l]{Best UMLS model}   & 60.20& \textbf{36.73} &  \textbf{36.73}\\
        
        &\makecell[l]{Best PrimeKG model}   &  59.18& 36.73 &  36.73\\
        \midrule
        \multirow{6}{*}{MedQA} &NFT               & 64.14 & 26.29 & 26.29 \\
        &RAG - UMLS               & 60.29& 26.29 & 26.29 \\
        &RAG - PrimeKG               & 47.80& 26.29 & 26.29  \\

        &Probsum SFT                    &  64.54 & 26.29 & 34.26\\

        &DDXPlus SFT                    & 61.75 & 25.10 & 26.69\\

        &\makecell[l]{Best UMLS model}   & \textbf{64.94}&  \textbf{26.29}& \textbf{26.29} \\

        &\makecell[l]{Best PrimeKG model}   & 64.94 & 26.29 & 26.29 \\
        \bottomrule \bottomrule
    \end{tabular}
    \vspace{-.1in}
    \caption{\small QA results on MedMCQA and MedQA dataset. Best performing training paradigm for each model from Table \ref{tab: probsum_ddxplus_rouge_cui} are evaluated alongside NFT models. The highest metric for each model is marked as bold, however the performance across different methods are similar and multiple methods achieve highest metric. }
    \label{tab: qa_task_result}
    \vspace{-.2in}
\end{table}

\begin{table*}[t!]
    \centering
    \resizebox{\textwidth}{!}{
    \begin{tabular}{ccccccccccccc}
    \toprule \toprule
        \multirow{3}{*}{PDSQI-9} & \multicolumn{3}{c}{NFT} & \multicolumn{3}{c}{Task SFT} & \multicolumn{3}{c}{Best UMLS model from Table \ref{tab: probsum_ddxplus_rouge_cui}} & \multicolumn{3}{c}{Best PrimeKG model from Table \ref{tab: probsum_ddxplus_rouge_cui}} \\
        \cmidrule(lr){2-4}\cmidrule(lr){5-7}\cmidrule(lr){8-10} \cmidrule(lr){11-13}
                                        & Qwn7B & Qwn8B & Gem7B & Qwn7B & Qwn8B & Gem7B & Qwn7B & Qwn8B & Gem7B & Qwn7B & Qwn8B & Gem7B \\
                                        \midrule
        Acc.            &  1.69  &  2.22  & 1.52 & 1.81$\uparrow$ &1.52 & 2.06$\uparrow$ & 1.62 & 2.12 & 1.68$\uparrow$ & 1.88$\uparrow$ & 1.80 & 1.32 \\
        Thorou.                    &  2.14  &  2.08  &  1.54 & 1.55& 1.32& 1.38& 2.16$\uparrow$ & 1.80 & 1.56$\uparrow$ & 2.06 & 1.84 & 1.44 \\
        Useful.                      &  2.46  &  2.48  &  1.72& 1.93& 1.42& 1.48& 2.38 & 2.12 & 1.78$\uparrow$ & 2.46 & 2.2 & 1.72 \\
        Org.                   &  4.12  &  2.70  &  2.94& 1.95&1.22& 1.42 & 3.64 & 2.50 & 3.00$\uparrow$ & 4.02 & 2.16 & 3.40$\uparrow$ \\
        Comp.               &  4.50  &  3.70  & 4.52&3.28&1.98& 2.58 & 4.32 & 3.56 & 4.62$\uparrow$ & 4.46 & 3.14 & 4.48 \\
        Succ.                   &  2.70  &  2.00  &  3.98&1.81&1.36& 1.86 & 2.54 & 2.10$\uparrow$ & 3.94 & 2.46 & 1.96 & 3.40 \\
        Synth.           &   3.00 &  2.32  &  0.76&1.45&0.64& 1.14$\uparrow$& 2.1 & 2.36$\uparrow$ & 0.60 & 2.74 & 1.86 & 1.42$\uparrow$ \\
        \midrule
        Avg.          & 2.94 & 2.50 & 2.42& 1.97& 1.35& 1.70 & 2.68 & 2.36 & 2.45$\uparrow$ & 2.86 & 2.13 & 2.45$\uparrow$ \\
        \bottomrule \bottomrule
    \end{tabular}
    }
    \vspace{-.1in}
    \caption{\small Best KG-trained LLM on ProbSum, evaluated by PDSQI-9 criteria, consisting of Accuracy (Extractive), Thoroughness, Usefulness, Organization, Comprehensibility, Succinctness, Synthesis/Abstraction metrics. Baselines include NFT models and Probsum SFT models. We compare the baselines with the best performing KG-trained models on the Probsum task, on Table \ref{tab: probsum_ddxplus_rouge_cui}. $\uparrow$ denotes an improvement over the baseline models. In case of UMLS KG, the best checkpoints for Qwn7B, Qwn8B, and Gem7B are GRPO, RM-R1, and SFT respectively. In case of PrimeKG, the best checkpoints for Qwn7B, Qwn8B, and Gem7B are GRPO, RM-R1, and RM-R1 respectively.}
    \label{tab: probsum_pdsqi9}
    \vspace{-.1in}
\end{table*}

\section{Task-level Results} 
\label{sec:task-level-results}
We first examine task-level performance across KG tasks and downstream benchmarks; §\ref{sec:gradient-analysis} then characterizes the same paradigms through optimization geometry.

\textit{Path judging vs. generation.} Figure~\ref{fig: task_heatmap} reports cross-task performance. Even without KG training, P@2 achieves high performance under both UMLS and PrimeKG (knowledge already encoded), whereas generative tasks such as \textsc{nhp} remain challenging, indicating a stronger reliance on explicit graph supervision. Models trained on path-judging tasks (P@10, P@2, PN@10) generalize well within the same family, but transfer poorly to generative tasks such as NHP and PC, and vice versa. This asymmetry suggests that path discrimination objective induces more task-specific biases that have limited generalizability.   

\textit{Multi-task robustness and training paradigm.} The multi-task SFT model shows the most consistent cross-task  performance across both KGs (Figure~\ref{fig: task_heatmap}),  improving robustness within judgment-based formulations but with limited generalization to generative tasks (NHP, PC). Across paradigms (Table~\ref{tab:training_paradigm}), SFT, GRPO, and RM-R1 achieve comparable in-domain performance, indicating that task-level accuracy alone cannot distinguish their reasoning behavior.

\textit{Downstream task performance.} On diagnosis prediction (Table~\ref{tab: probsum_ddxplus_rouge_cui}), task-specific SFT achieves the highest in-domain performance on ProbSum and DDXPlus; reward-trained models consistently outperform NFT but not task-specific SFT. On medical QA (Table~\ref{tab: qa_task_result}), results are mixed: task-specific SFT wins on MedMCQA, while KG-trained models match or exceed it on MedQA. Diagnosis-prediction SFT checkpoints show intermediate QA performance, suggesting partial transfer.

\textit{Clinical reasoning quality.}  On PDSQI-9 (Table~\ref{tab: probsum_pdsqi9}), KG-trained models improve higher-level reasoning dimensions (organization, comprehensibility, synthesis) on Qwn7B and Gem7B, while showing no consistent gains on extractive accuracy or thoroughness. This pattern, structural rather than surface-level gains, motivates the gradient-level analysis in §\ref{sec:gradient-analysis}.

\begin{figure*}[ht!]
    \centering
    \includegraphics[width=\textwidth]{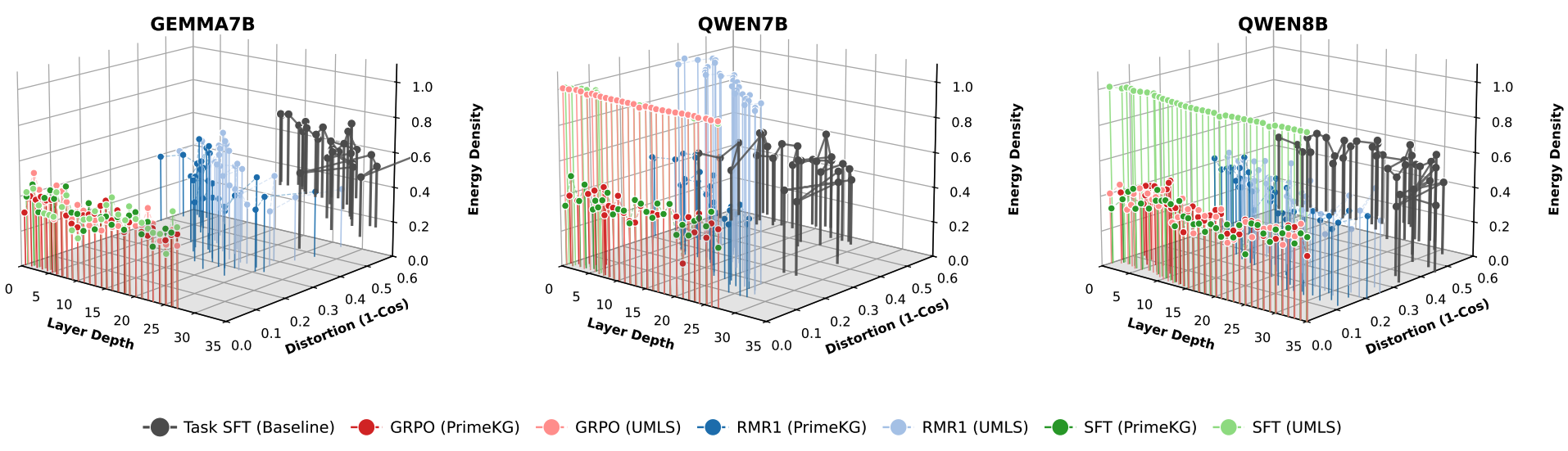}
    \vspace{-.35in}
    \caption{Gradient analysis across Models and training paradigms. We map the layer-wise gradients using Layer Depth (X), Gradient Distortion (Y), and \textbf{Gradient Intervention Density} (Z).}    
    \label{fig:gradient_analysis}
    \vspace{-0.15in}
\end{figure*}

\section{Gradient Analysis}
The task-level patterns in \S\ref{sec:task-level-results} raise a deeper question: what has each training paradigm actually done to the pretrained model? To answer this, we analyze the geometry of layer-wise parameter updates relative to the non-finetuned baseline (NFT) and characterize this comparison along two dimensions: \textit{direction} (cosine 
alignment) and \textit{magnitude} (norm ratio). For each fine-tuned model $A$ and the NFT baseline model B, we calculate the normalized gradients of the attention parameters at each layer ($\mathbf{g}_A$ and $\mathbf{g}_B$ respectively). For LoRA-trained models, we merge the adapter weights with the base model weights before computing the gradients.
\label{sec:gradient-analysis}
\subsection{Diagnostic Framework}

First, we use cosine similarity to measure the geometric alignment of layer-wise deltas $\mathbf{g}_A$ and $\mathbf{g}_B$ from two different models~\cite{yu2020gradient, chen2018gradnorm}:
\begin{equation}
    \text{Cos}(\mathbf{g}_A, \mathbf{g}_B) = 
    \frac{\mathbf{g}_A \cdot \mathbf{g}_B}{\|\mathbf{g}_A\| \|\mathbf{g}_B\|}.
\end{equation}
where lower cosine similarity indicates larger directional divergences. 

Second, we use \textit{energy shift} to measure the relative intensity of updates. Since gradient norms reflect loss landscape geometry~\cite{he2019asymmetric} and local convergence behavior~\cite{damian2023smoothing}, we quantify this using the logarithmic difference in $L_2$ norms:
\begin{equation}
\small 
    \text{Energy Shift}(\mathbf{g}_A, \mathbf{g}_B) = \log(\|\mathbf{g}_A\|) - \log(\|\mathbf{g}_B\|)
\end{equation}
A positive shift indicates that model $A$ exhibits more aggressive optimization energy in that layer compared to model $B$. 

Raw direction and magnitude are continuous quantities sensitive to outliers in any single layer. More importantly, they do not directly answer the question we care about: \textit{which components in the pretrained model has the optimizer actually changed?} We therefore treat optimization as a discrete intervention process. For each layer, we define an \textsc{Intervention Event} as a parameter delta whose $L_2$-norm shift exceeds a strict tolerance $\delta = 10^{-9}$, chosen to separate optimization updates from floating-point noise. (Appendix~\ref{appendix: gradient_threshold_analysis} presents a sensitivity analysis of $\delta \in \{ [10^{-6}, 10^{-10}]\}$, where the trends are stable across $\delta$). Under this definition, a component is either\textit{touched} or \textit{mathematically invariant}. 

We introduce \textbf{Gradient Intervention Density (GID)}, the fraction of layer components classified as touched. GID provides a single scalar per layer that captures the \textit{scope}, not the \textit{magnitude}, of optimizer activity: the structural property that, we argue, governs how training reshapes pretrained representations. We complement GID with \textbf{Gradient Distortion (GD)} ($1 - \text{Cos}$) for directional misalignment. Unlike gradient-conflict analyses in multi-task learning~\cite{yu2020gradient}, which characterize how multiple objectives interfere during training, GID isolates the discrete footprint of a single training objective relative to the pretrained anchor, providing a direct diagnostic comparable across different training checkpoints. 


\begin{table}[t!]
    \centering
    \scriptsize
    \setlength{\tabcolsep}{2pt}
    \begin{tabular}{llcccc}
        \toprule \toprule
        && \multicolumn{2}{c}{KG SFT} & \multicolumn{2}{c}{ProbSum SFT} \\
        \cmidrule(lr){3-4} \cmidrule(lr){5-6}
        Task &$\lambda$ & Rouge-L & CUI-F & Rouge-L & CUI-F \\
        \midrule
        \multirow{2}{*}{Probsum} &0.0 & 18.88 $\pm$ 0.03 & 24.00 $\pm$ 0.00 & 22.92 $\pm$ 0.01 & 25.87 $\pm$ 0.00 \\
        &0.1& 18.94 $\pm$ 0.02 & 22.66 $\pm$ 0.00 & 19.48 $\pm$ 0.03 & 24.88 $\pm$ 0.00\\
        \midrule
        && \multicolumn{4}{c}{Out-of-distribution test set} \\  
        \midrule
        && \multicolumn{2}{c}{Accuracy} & \multicolumn{2}{c}{Accuracy} \\
        \multirow{2}{*}{MedQA}& 0.0 & \multicolumn{2}{c}{64.14 $\pm$ 1.17} & \multicolumn{2}{c}{64.54 $\pm$ 0.65}\\
        & 0.1 & \multicolumn{2}{c}{64.01 $\pm$ 1.17} & \multicolumn{2}{c}{63.61 $\pm$ 0.18}\\
        \midrule
        \multirow{2}{*}{MedMCQA}& 0.0 & \multicolumn{2}{c}{57.82 $\pm$ 2.54} & \multicolumn{2}{c}{59.18 $\pm$ 0.00}\\
        & 0.1 & \multicolumn{2}{c}{57.14 $\pm$ 0.00} & \multicolumn{2}{c}{57.82 $\pm$ 0.96}\\
        \bottomrule \bottomrule
    \end{tabular}
    \vspace{-.1in}
    \caption{\small Ablation study on Qwen2.5-7B-Instruct model. We report the average ROUGE-L score $\pm$ standard deviation (sd) across three runs on the Probsum diagnosis task. Furthermore, we also report mean accuracy scores $\pm$ sd across three runs on the MedQA and MedMCQA dataset.}
    \label{table:ablation-results}
    \vspace{-.1in}
\end{table}


\subsection{Results}
\label{sec:sec7-results}

\begin{figure}[h!] 
   \centering
   \includegraphics[height=0.6\columnwidth]{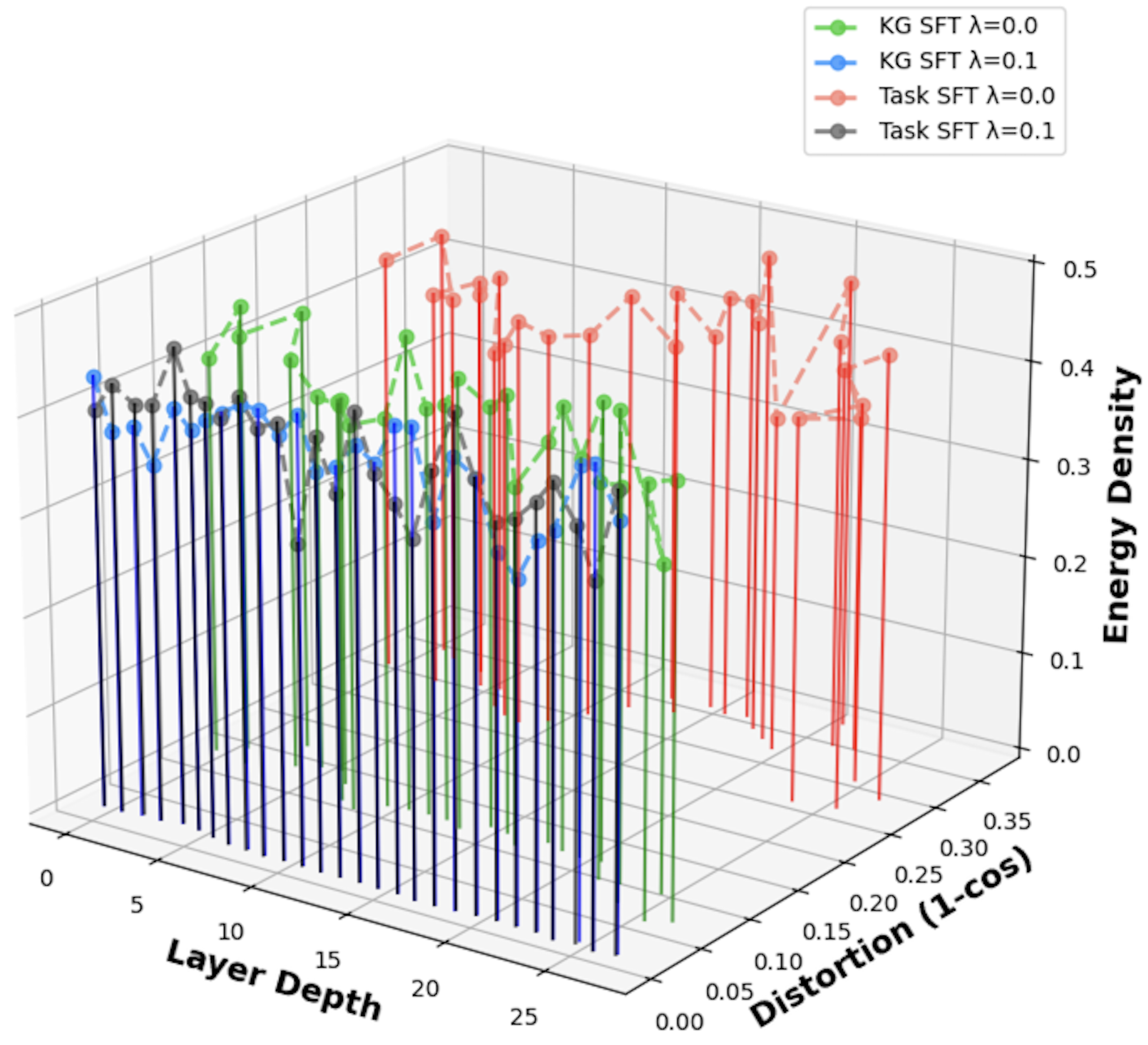}
    \caption{\small Gradient analysis across Qwn7B models in our ablation study.}
    \label{fig: gradient-ablation}
    \vspace{-.2in}
\end{figure}

Figure~\ref{fig:gradient_analysis} visualizes layer-wise training dynamics by mapping layer depth (X), gradient distortion (Y), and Gradient Intervention Density (Z) relative to the non-finetuned baseline. Task-specific SFT (gray) modifies almost every component at every layer with high distortion (GD near 1.0). KG-guided methods (GRPO and KG-SFT, in green/red shades) modify conservatively (GD well below 1.0): the sparse, surgical updates we predicted. RM-R1 (blue shades) sits in between, consistent with its lack of explicit KL regularization. Even in cases where GID for KG guided models are high, the GD is still low.

This ordering holds across all three base models (Gemma-7B, Qwen2.5-7B, Qwen3-8B), suggesting the pattern reflects the  training paradigm rather than the model. We term this sparse  regime \underline{\textit{surgical alignment}}: parameter updates that touch only what is needed while leaving most of the pretrained model intact. 

The paradigms with the sparsest GID profiles (KG-SFT and GRPO)  are the same ones that improved Qwn7B and Gem7B on higher-order PDSQI-9 dimensions in \S\ref{sec:task-level-results}. Task SFT, despite winning in-domain ROUGE-L, did not improve those dimensions. In short: \textit{where} the optimizer puts its updates matters more for clinical reasoning quality than \textit{how much} it updates. 

\subsection{Ablation: objective vs. KL divergence}
\label{sec:ablation}

Is the sparse update footprint of surgical alignment driven by the KG-judgment objective, by KL regularization, or by both? Without disentangling these factors, the §\ref{sec:sec7-results} finding could collapse into a familiar fact: KL regularization always produces small updates. We run a controlled $2\times2$ ablation to determine which factor matters. 

We compare two objectives, Task-specific SFT (on ProbSum) and Multi-task KG-judgment SFT, each trained at KL strength $\lambda \in \{0, 0.1\}$. All four cells use Qwn7B, matched training data size (multi-task KG SFT is subsampled to ProbSum size), identical LoRA configuration, fixed learning rate (no per-cell tuning), equal training steps, and 3 random seeds. 
Table~\ref{table:ablation-results} reports task-level performance across the four cells; gradient-level results are in figure~\ref{fig: gradient-ablation}. Three patterns emerge:  

1) \textit{Task-specific supervision dominates in-domain.} At $\lambda = 0$, ProbSum SFT achieves the highest in-domain 
ROUGE-L on ProbSum (22.92 vs.~KG SFT's 18.88), as expected.  

2) \textit{ProbSum SFT is sensitive to KL.} Adding KL drops 
its performance on every metric, even for out-of-distribution test set: ProbSum ROUGE-L 
$22.92 \to 19.48$, MedQA $64.54 \to 63.61$, MedMCQA 
$59.18 \to 57.82$.

3) \textit{KG SFT is robust to KL.} It is essentially flat 
across $\lambda$ on all three evaluations (ProbSum 
$18.88 \to 18.94$; MedQA $64.14 \to 64.01$; MedMCQA 
$57.82 \to 57.14$). Adding KL does not meaningfully change 
KG SFT's behavior.

Figure~\ref{fig: gradient-ablation} confirms why: KG-judgment SFT already produces a sparse update footprint at $\lambda = 0$, before any KL regularization is applied. The optimizer touches only a subset of components, so adding KL has little additional effect. ProbSum SFT, by contrast, modifies the parameter set broadly at $\lambda = 0$, and KL must subsequently constrain it (at the cost of in-domain accuracy).

The objective alone contributes to sparsity, which KL just stacks on top off. This further confirms that surgical alignment is \textit{a property of the KG-judgment objective}, not an artifact of regularization. Practically, this means objective selection is the primary lever for surgical updates, where task SFT could fail at preserving the pretrained model's broader clinical reasoning even when winning on the trained benchmark.



\section{Conclusion}
Through Gradient Intervention Density (GID) and Gradient Distortion (GD), we identify 
\textit{surgical alignment} as a property of judgment-based KG 
training: sparse, localized parameter updates that preserve the 
pretrained model's broader reasoning. Optimization geometry 
should be a first-class evaluation dimension alongside task 
accuracy for clinical KG-LLM integration. 


\section{Acknowledgments}
This work is supported by U.S. National Library of Medicine, National Institute of Health, under award number R00LM014308.
\section{Limitations}
While our work focuses on the systematic evaluation of LLMs across different task formulations, training paradigms, and knowledge graphs (KGs), this evaluation can be further expanded to better understand training dynamics across diverse domains. Furthermore, we limit our study to three commonly used LLMs; however, this work can be extended to provide a broader evaluation of additional LLMs. Finally, while we introduce a mechanistic perspective for understanding KG–LLM integration through GID, we do not propose a novel RL training framework aimed at improving performance metrics. 
\section{Ethical Considerations}
This study uses the ProbSum dataset, which is built using the MIMIC-III dataset which is a deidentified clinical datasets containing no personally identifiable information. For generating reasoning traces, we used Azure GPT-o3-mini and for the PDSQI-9 evaluations, we use Azure GPT-5-mini (ICC $\geq$ 0.8) as the LLM-as-judge to score the ProbSum test set. Azure is a HIPAA-compliant environment and therefore complies with the MIMIC Data Use Agreement.
Our motivation is to investigate and enhance the capabilities of current LLMs in the context of diagnostic reasoning. While AI tools have substantial potential to support healthcare professionals, existing LLMs exhibit limitations that must be addressed to enable their safe and effective adoption in clinical settings. Key challenges include mitigating hallucinations, preserving patient privacy, and overcoming other factors that may compromise reliability and trustworthiness.

\bibliography{custom}

\appendix

\section{Training Details} \label{training_details}
For all SFT trainings, we apply Kullback-Leibler (KL) Divergence regularization~\cite{6639201}. This limits the divergence of internal state representation of the finetuned models from the non-finetuned model. For each task formulation, we perform hyperparameter tuning for 10 trials using the Optuna framework~\cite{akiba2019optunanextgenerationhyperparameteroptimization} and select the KL-divergence hyperparameter between 0.01 and 1. Moreover, to finetune models efficiently, we utilize Low-Rank Adaptation (LoRA)~\cite{hu2021loralowrankadaptationlarge} technique and apply 4-bit NF4 quantization~\cite{dettmers2023qloraefficientfinetuningquantized}. We also limit our training to only the attention layers and apply LoRA with rank 16. For GRPO training, we sample 6 candidate generations with maximum new token length of 256 from the model and use 0.001 as the KL-divergence hyperparameter. For the GRPO stage in the RM-R1 paradigm, we sample 6 candidate generations with a maximum new token length of 512 to work within the available hardware. All the trainings were performed on a Dell server with two Nvidia H100 GPUs. We also apply early stopping with \textsc{patience} of 2 for all our trainings.

\section{Evaluation setup} \label{appendix: evaluation_setup_additional}
We define the Precision and Recall to evaluate concept grounded F1 score (CUI-F) as:
\begin{equation}\label{eq:cui-f}
\scriptsize
\begin{split}
\text{Precision}
= \frac{|Predicted \cap Gold|}{|Predicted|}
\quad
\text{Recall}
= \frac{|Predicted \cap Gold|}{|Gold|}
\end{split}
\end{equation}
\vspace{-0.15in}

Where Predicted represents the set of KG concepts in model's prediction and Gold reflects the set of KG concepts in ground truth label.



\section{Detailed training dynamics}
\label{appendix:detailed_gradient_results}

\begin{figure*}[p]
    \centering
    \setlength{\tabcolsep}{1pt}
    \renewcommand{\arraystretch}{0.05}
    \scriptsize
    \resizebox{0.85\textwidth}{!}{%
    \begin{tabular}{@{}lcccccccc@{}}
        & {\scriptsize Baseline} && {\scriptsize MT SFT (UMLS)} && {\scriptsize MT GRPO (UMLS)} && {\scriptsize MT RMR1 (UMLS)} \\[2pt]
        
        \rotatebox{90}{\scriptsize Gemma-7B} &
        \includegraphics[width=0.23\textwidth]{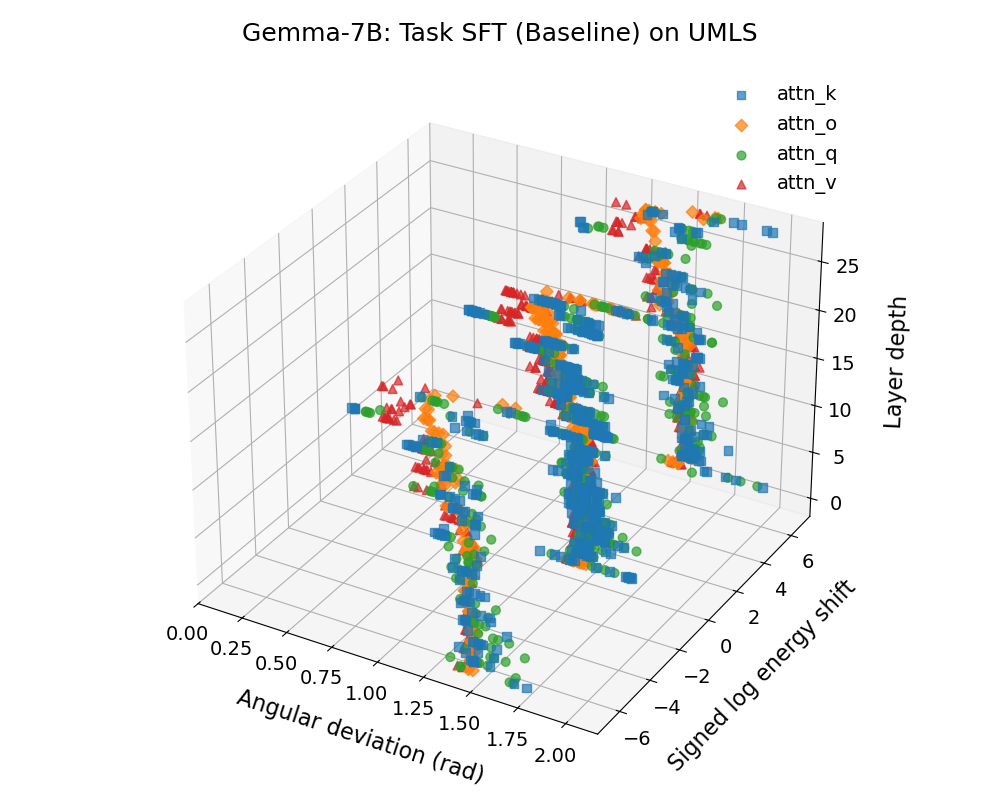} &&
        \includegraphics[width=0.23\textwidth]{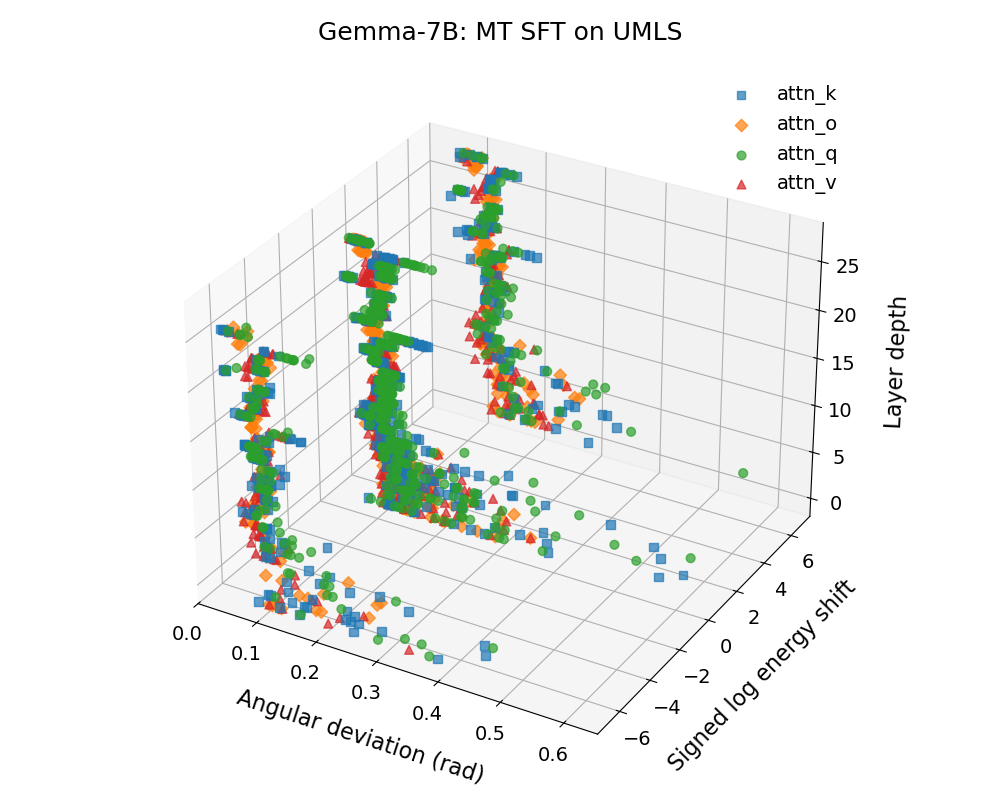} &&
        \includegraphics[width=0.23\textwidth]{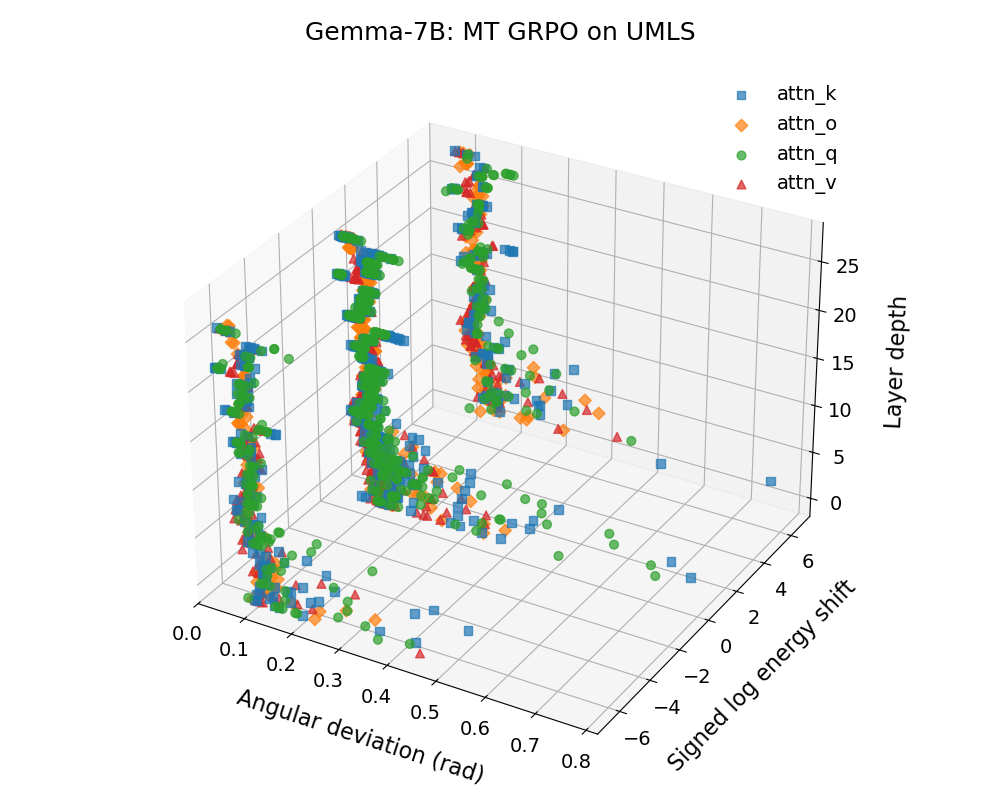} &&
        \includegraphics[width=0.23\textwidth]{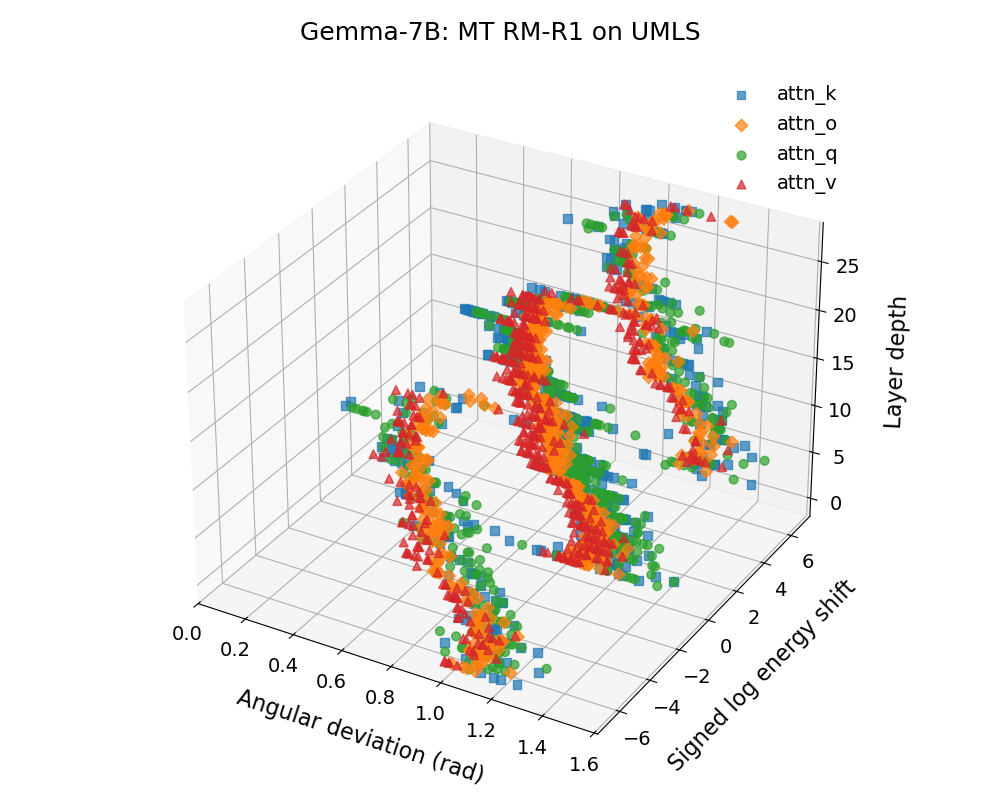} \\[2pt]
        
        \rotatebox{90}{\scriptsize Qwen2.5-7B} &
        \includegraphics[width=0.23\textwidth]{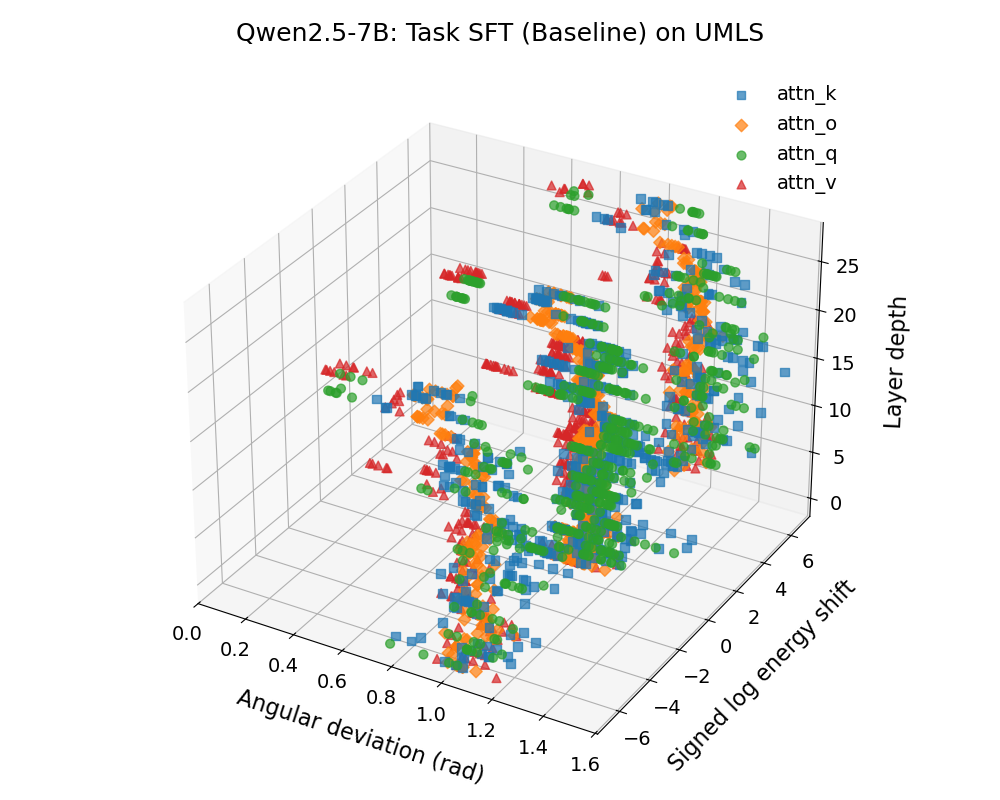} &&
        \includegraphics[width=0.23\textwidth]{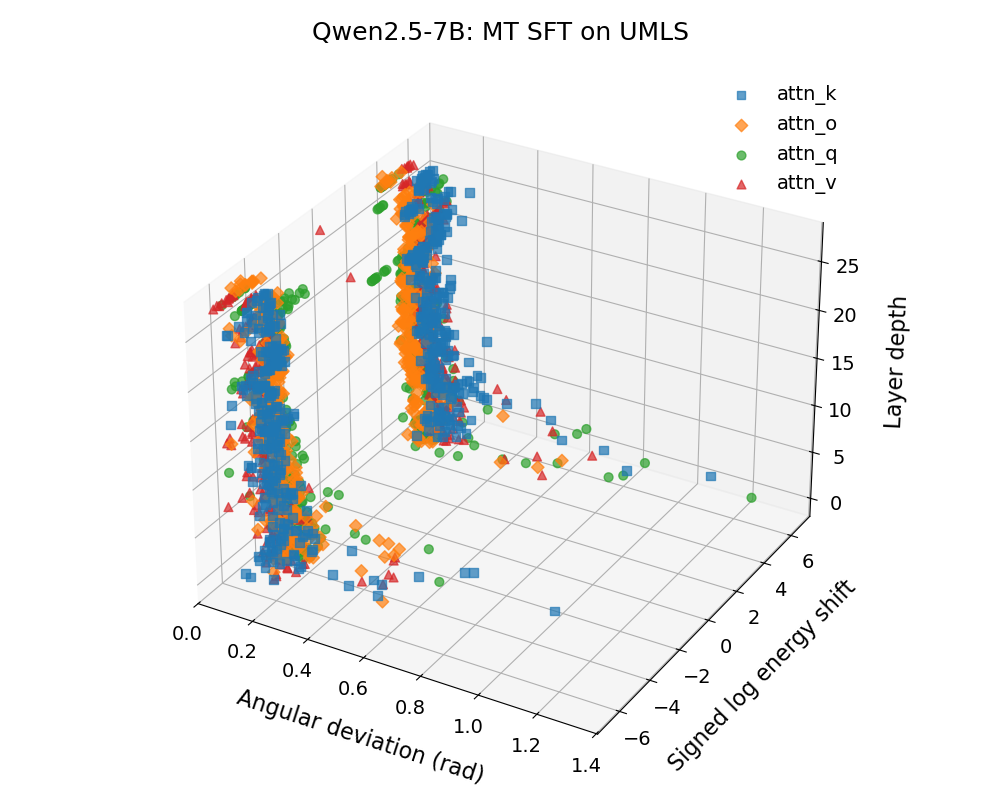} &&
        \includegraphics[width=0.23\textwidth]{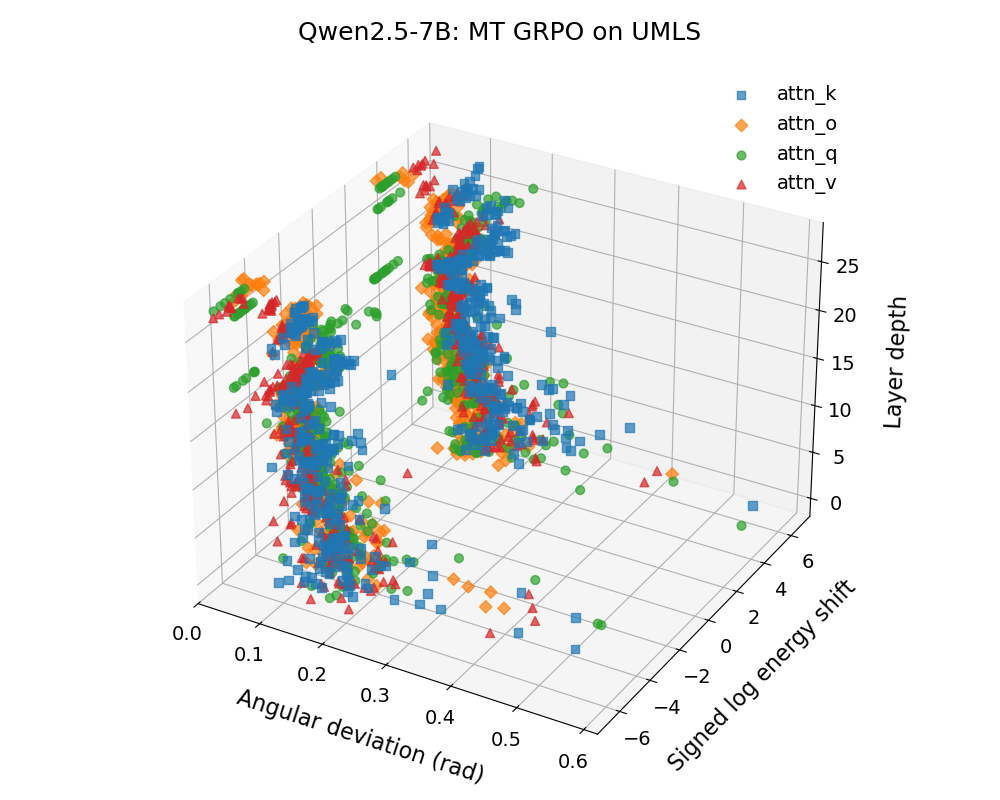} &&
        \includegraphics[width=0.23\textwidth]{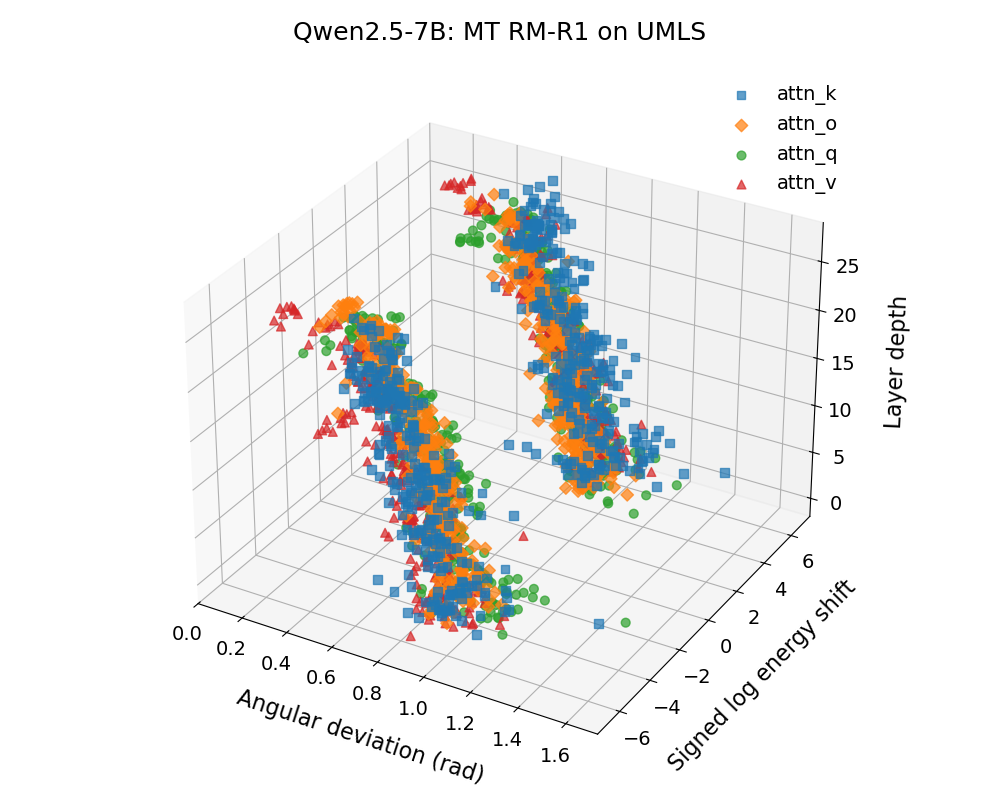} \\[2pt]
        
        \rotatebox{90}{\scriptsize Qwen3-8B} &
        \includegraphics[width=0.23\textwidth]{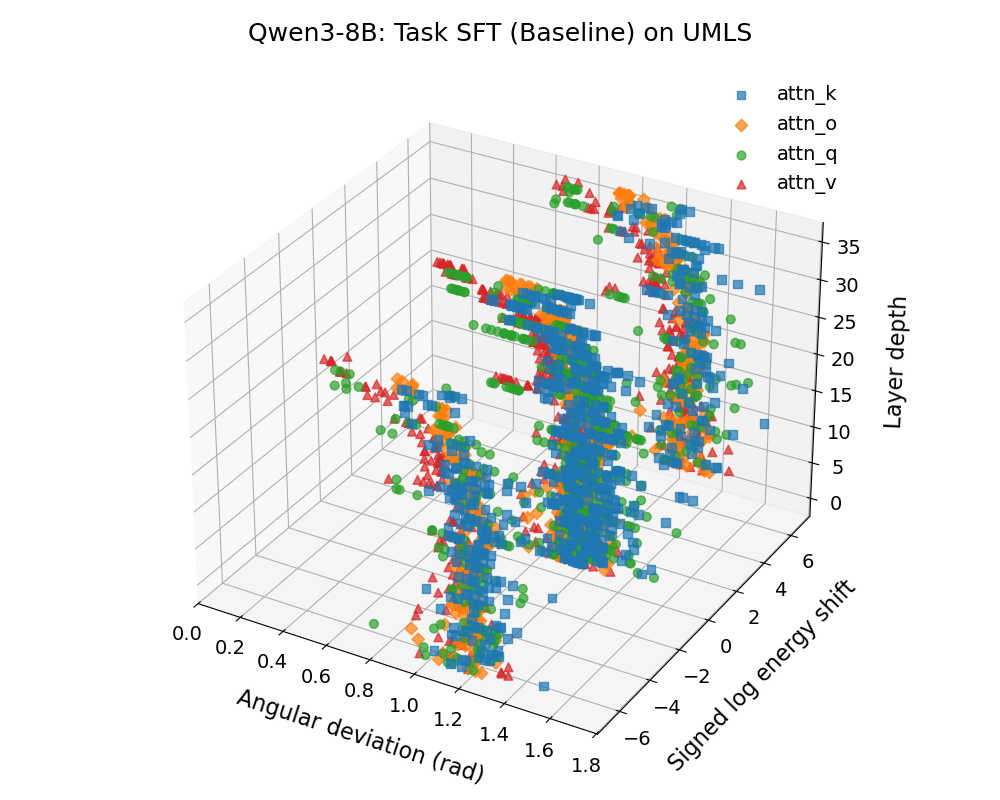} &&
        \includegraphics[width=0.23\textwidth]{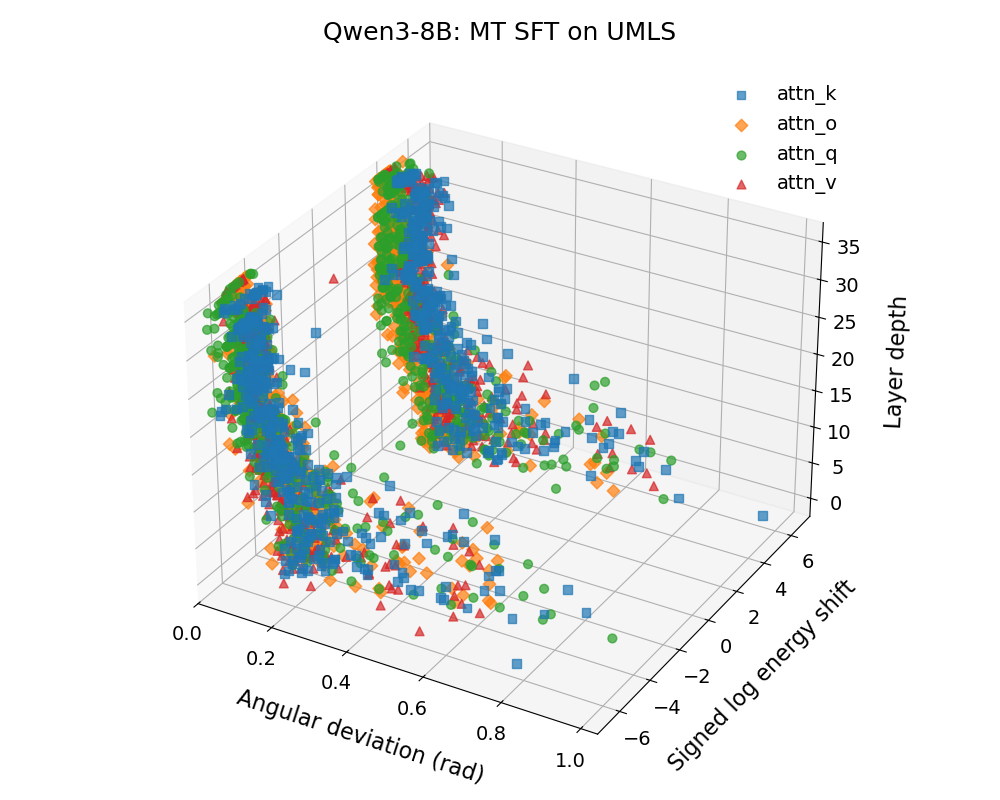} &&
        \includegraphics[width=0.23\textwidth]{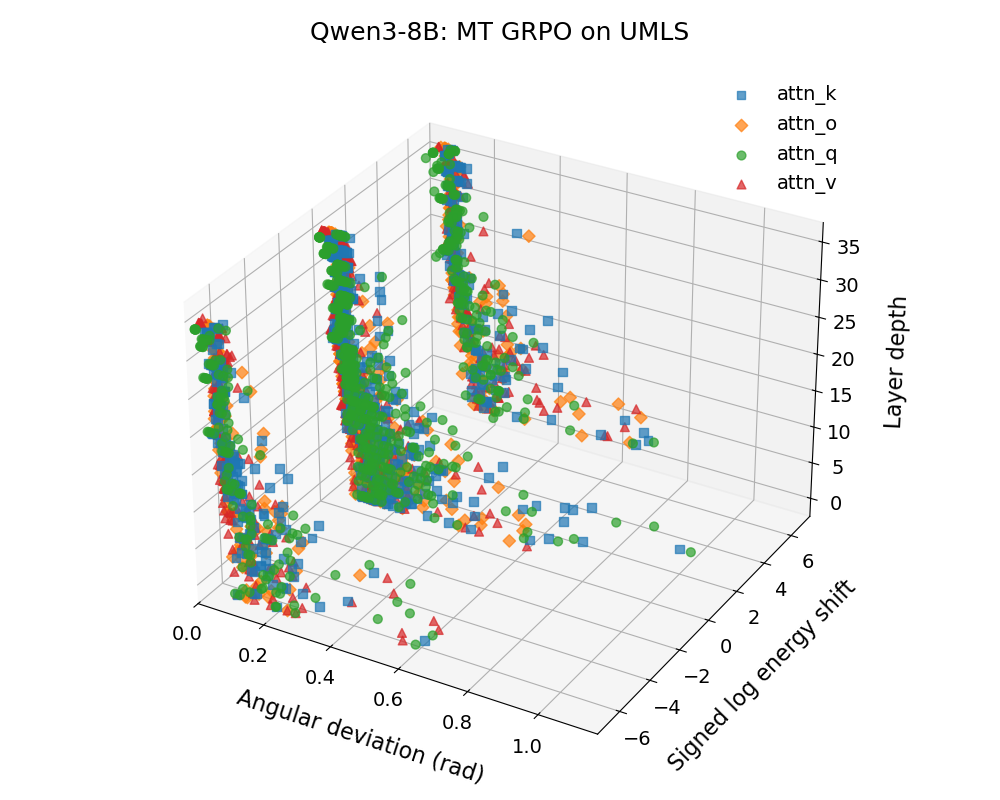} &&
        \includegraphics[width=0.23\textwidth]{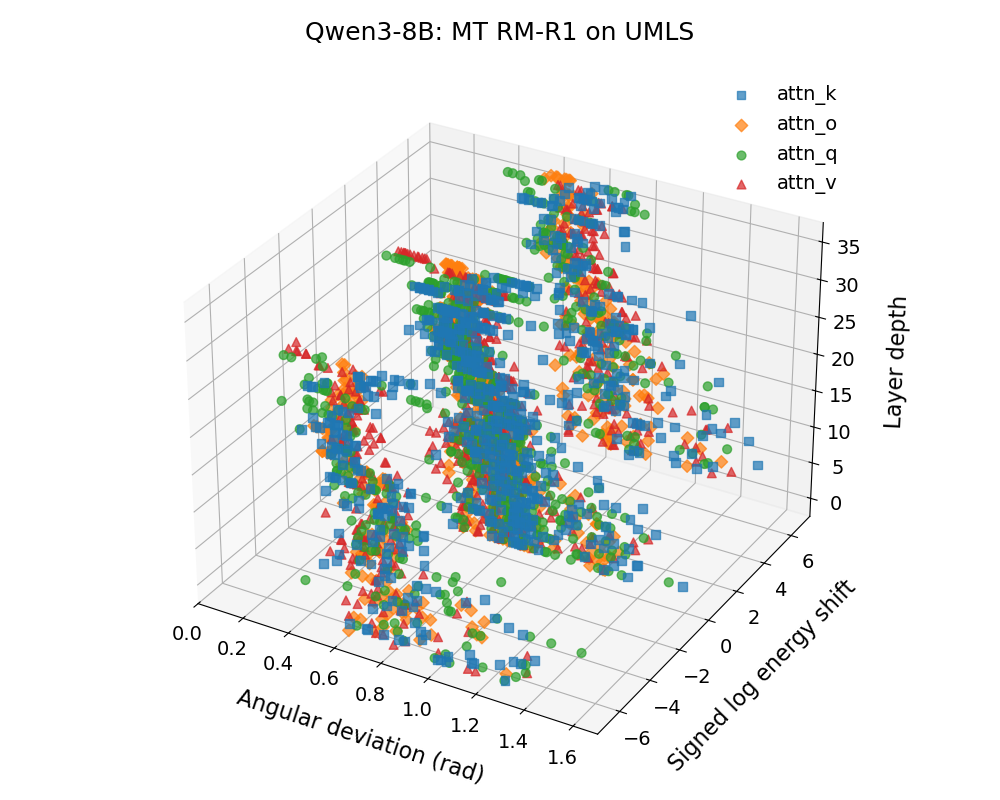} \\
    \end{tabular}
    }
    \caption{\small \textbf{Gradient-level optimization dynamics.} Layer-wise analysis for Baseline and UMLS-based methods across three models.}
    \label{fig:gradient_analysis_part1}
\end{figure*}

\begin{figure*}[p]
    \centering
    \setlength{\tabcolsep}{1pt}
    \renewcommand{\arraystretch}{0.05}
    \scriptsize
    \resizebox{0.85\textwidth}{!}{%
    \begin{tabular}{@{}lcccccccc@{}}
        & {\scriptsize Baseline} && {\scriptsize MT SFT (PrimeKG)} && {\scriptsize MT GRPO (PrimeKG)} && {\scriptsize MT RMR1 (PrimeKG)} \\[2pt]
        
        \rotatebox{90}{\scriptsize Gemma-7B} &
        \includegraphics[width=0.23\textwidth]{sections/figures/plots_3d_png/nft_vs_tasksft_umls_gemma7b.png} &&
        \includegraphics[width=0.23\textwidth]{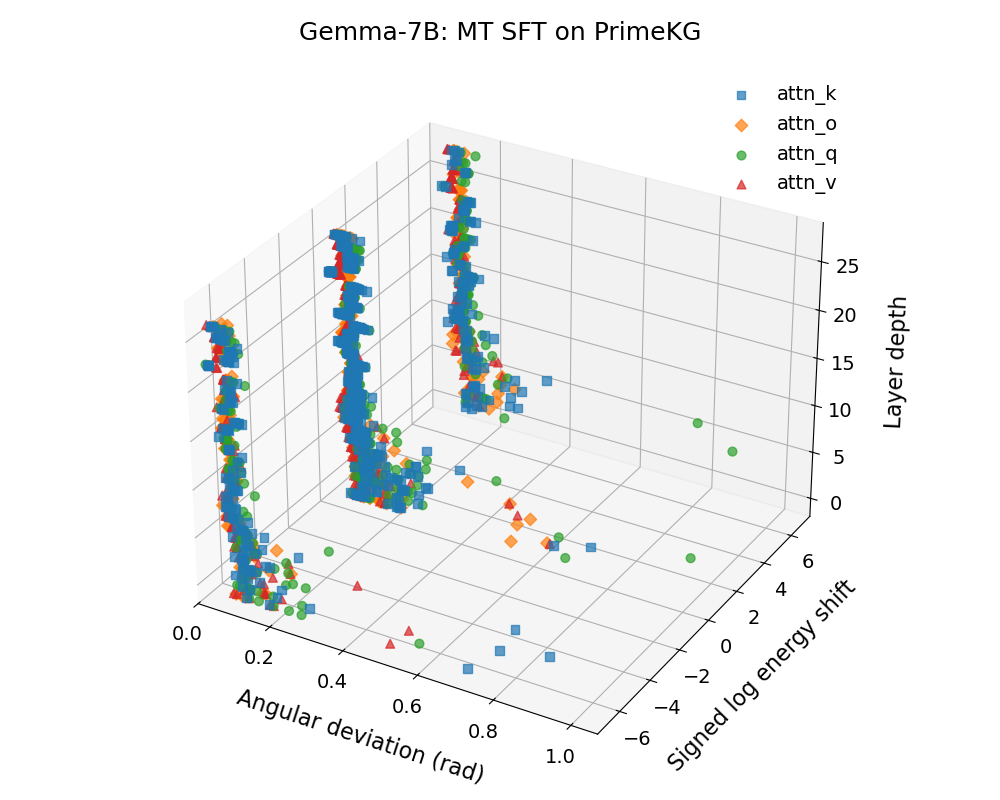} &&
        \includegraphics[width=0.23\textwidth]{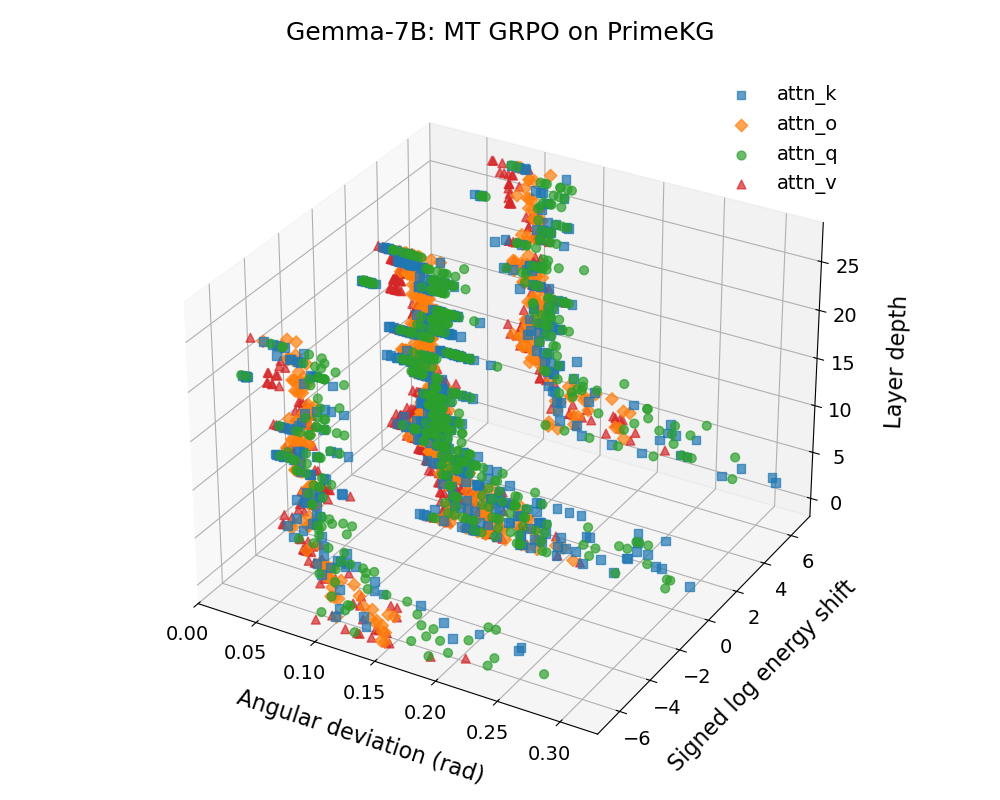} &&
        \includegraphics[width=0.23\textwidth]{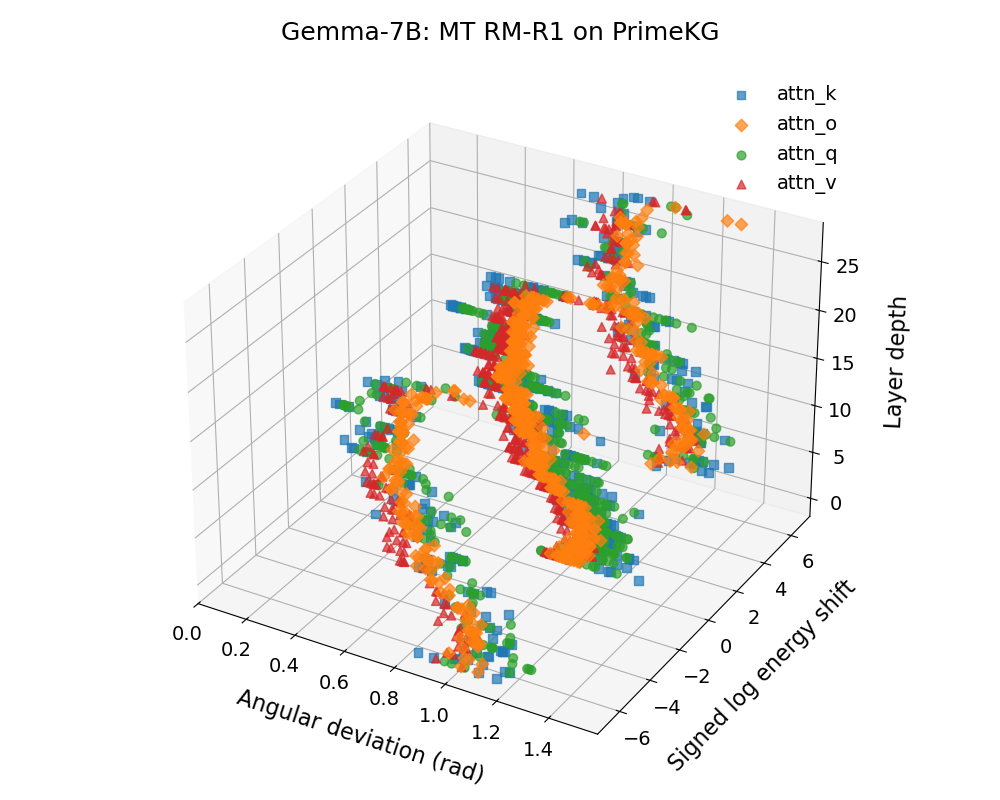} \\[2pt]
        
        \rotatebox{90}{\scriptsize Qwen2.5-7B} &
        \includegraphics[width=0.23\textwidth]{sections/figures/plots_3d_png/nft_vs_tasksft_umls_qwen7b.png} &&
        \includegraphics[width=0.23\textwidth]{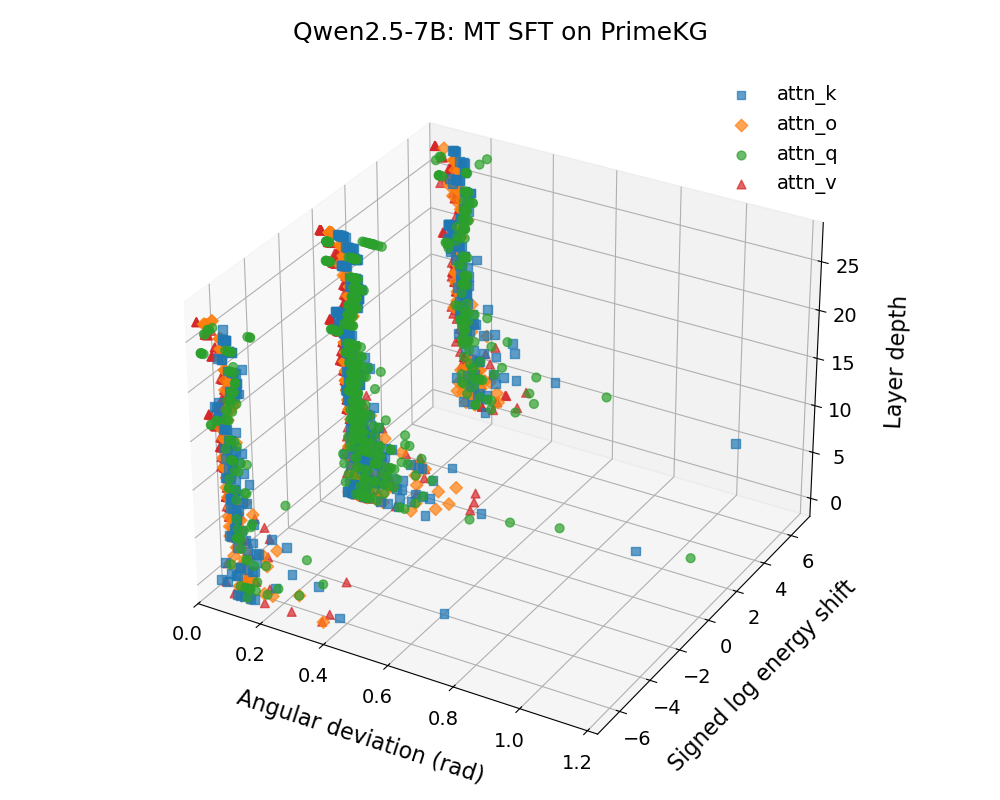} &&
        \includegraphics[width=0.23\textwidth]{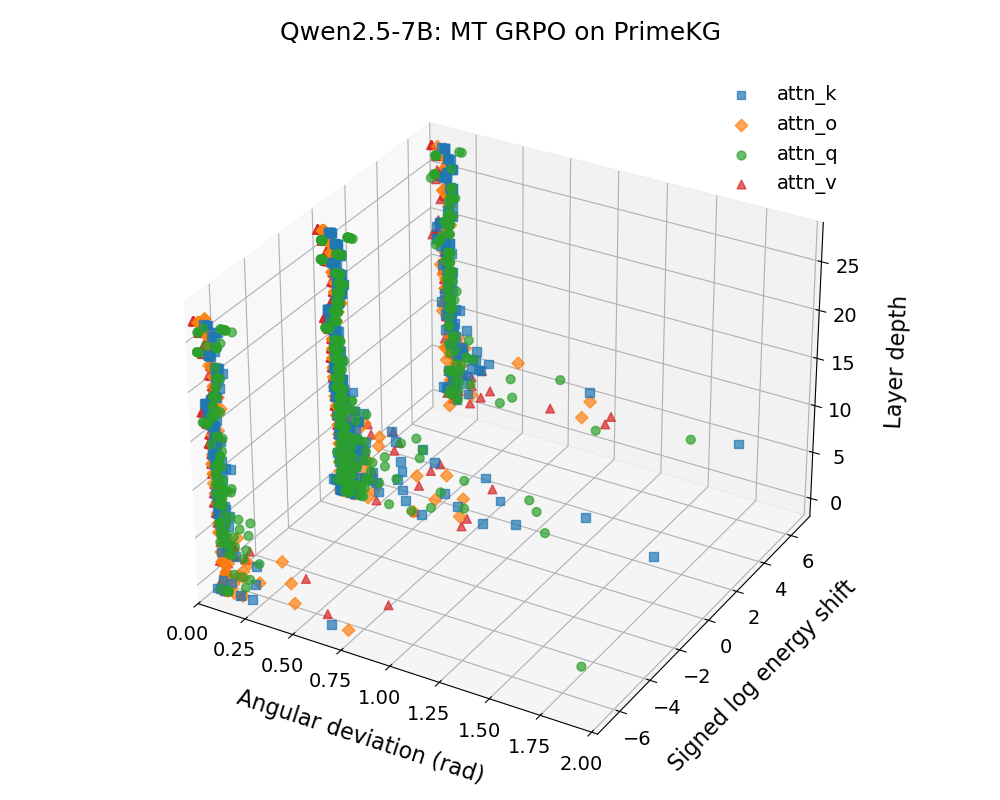} &&
        \includegraphics[width=0.23\textwidth]{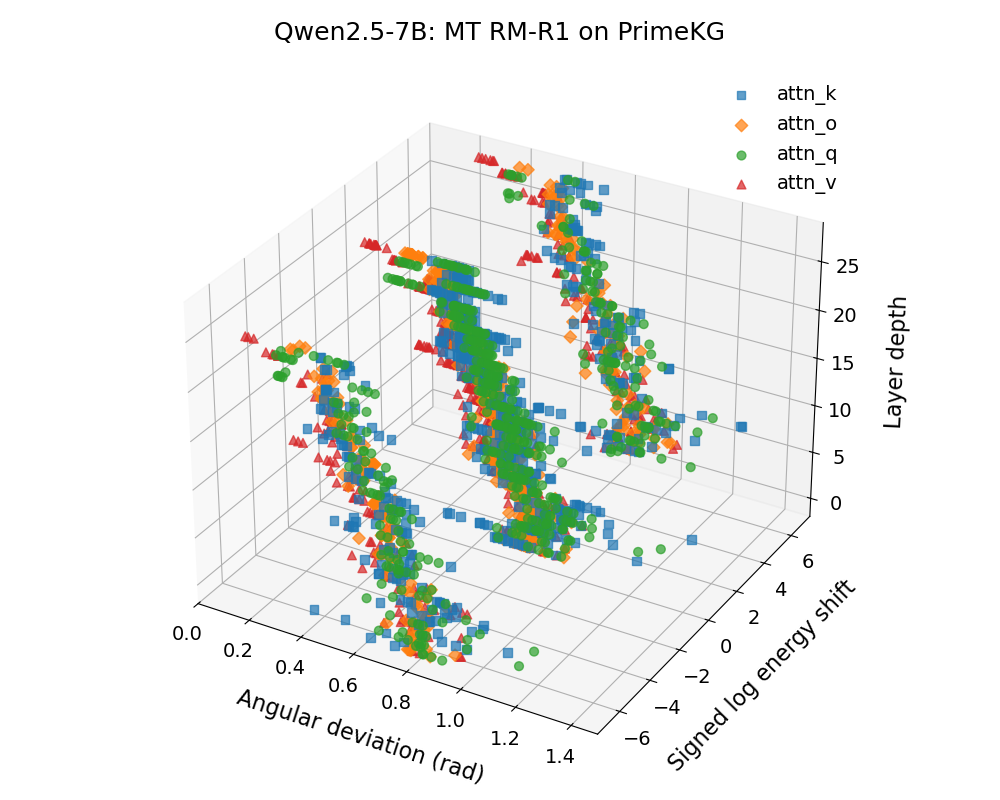} \\[2pt]
        
        \rotatebox{90}{\scriptsize Qwen3-8B} &
        \includegraphics[width=0.23\textwidth]{sections/figures/plots_3d_png/nft_vs_tasksft_umls_qwen8b.png} &&
        \includegraphics[width=0.23\textwidth]{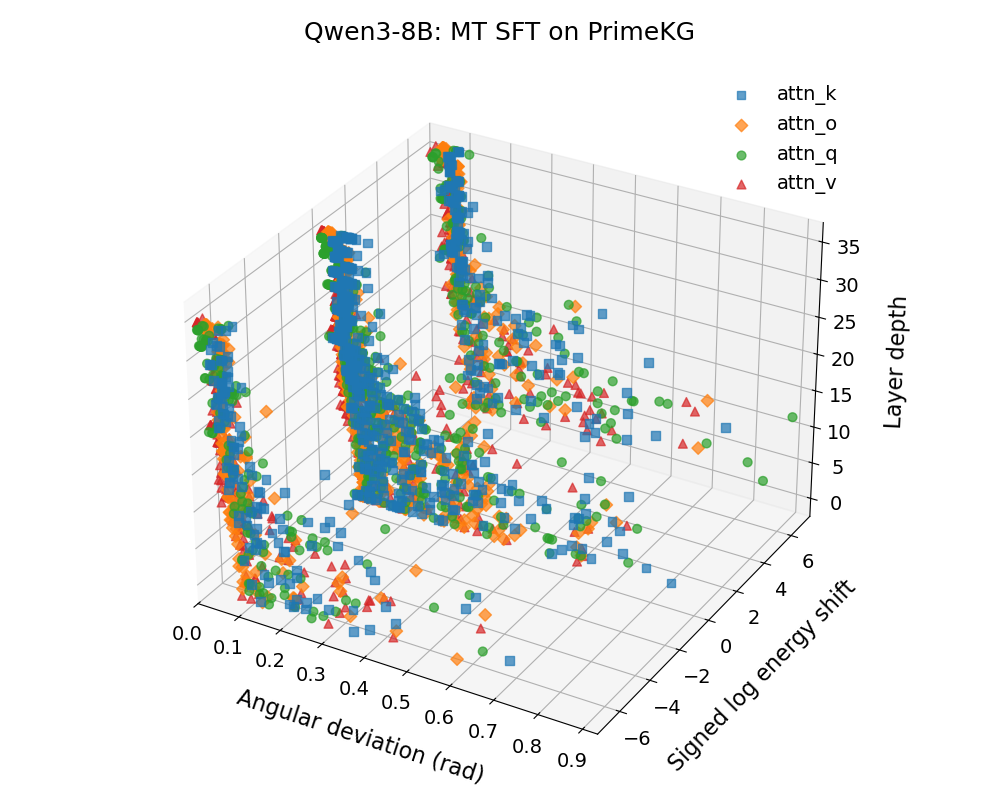} &&
        \includegraphics[width=0.23\textwidth]{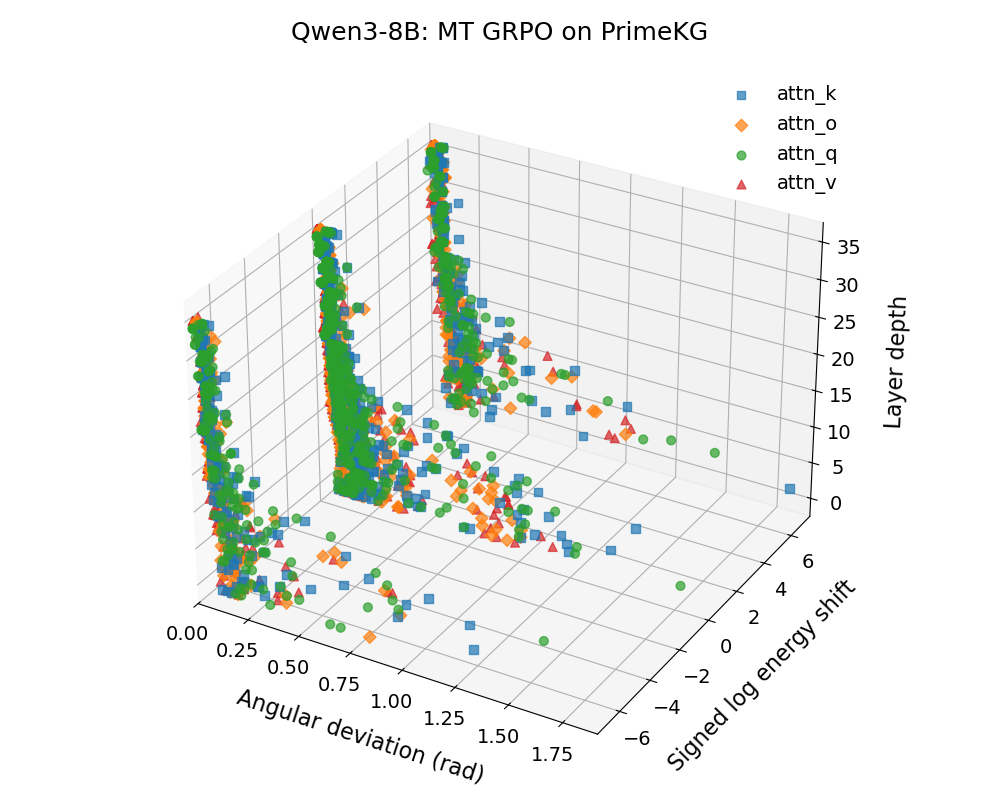} &&
        \includegraphics[width=0.23\textwidth]{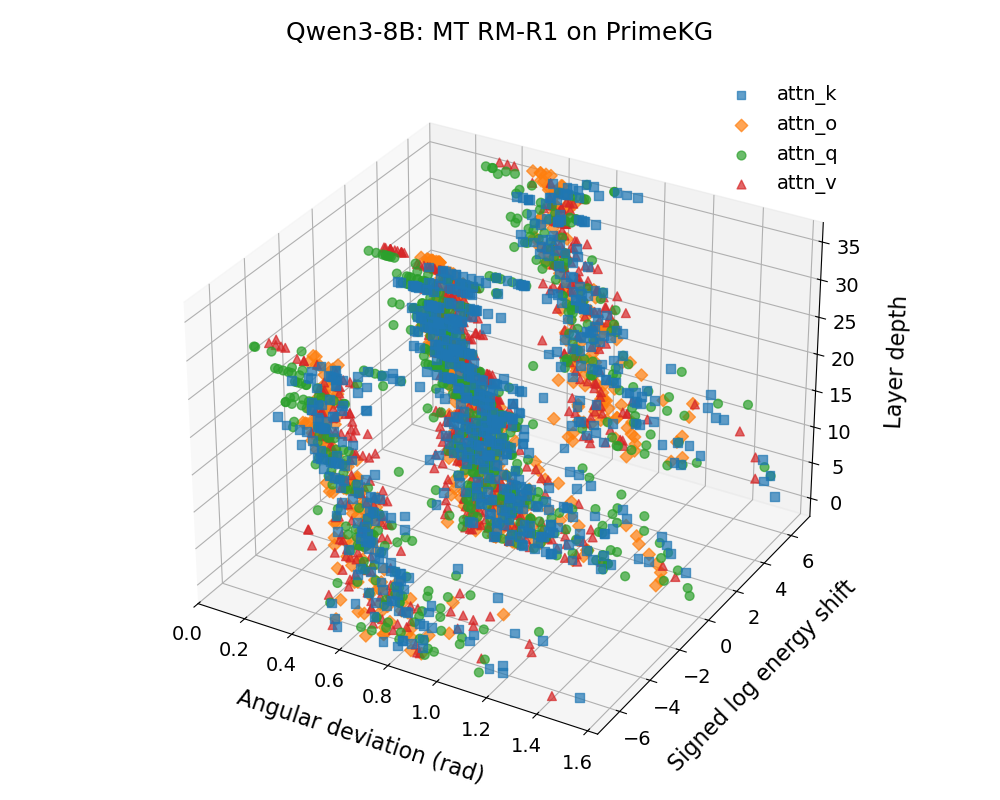} \\
    \end{tabular}
    }
    \caption{\small \textbf{Gradient-level optimization dynamics.} Layer-wise analysis for Baseline and PrimeKG-based methods across three models.}
    \label{fig:gradient_analysis_part2}
\end{figure*}

\begin{table*}[h]
\centering
\resizebox{0.95\textwidth}{!}{%
\begin{tabular}{lll cccc cccc cccc }
\toprule
\multirow{2}{*}{\textbf{KG}} & \multirow{2}{*}{\textbf{Method}} & \multirow{2}{*}{\textbf{Met.}} & \multicolumn{4}{c}{\textbf{Qwen2.5-7B}} & \multicolumn{4}{c}{\textbf{Qwen3-8B}} & \multicolumn{4}{c}{\textbf{Gemma-7B}} \\
\cmidrule(lr){4-7} \cmidrule(lr){8-11} \cmidrule(lr){12-15} 
& & & Q & K & V & O & Q & K & V & O & Q & K & V & O \\
\midrule
\multirow{4}{*}{\textbf{Baseline}} & \multirow{4}{*}{Task SFT} & \texttt{Cos} & 0.4274 & 0.4253 & 0.5275 & 0.4544 & 0.3784 & 0.3424 & 0.4366 & 0.3597 & 0.1137 & 0.1224 & 0.1929 & 0.1470 \\
 &  & \texttt{Mse} & 8.9e-08 & 6.3e-07 & 5.1e-07 & 8.5e-08 & 7.4e-08 & 3.1e-07 & 2.7e-07 & 7.6e-08 & 1.4e-07 & 1.4e-07 & 1.3e-07 & 1.4e-07 \\
 &  & \texttt{Egy} & -5.2e-09 & 1.9e-09 & 1.2e-09 & 8.5e-10 & -2.5e-10 & 2.3e-09 & -1.3e-09 & -8.3e-11 & -1.8e-09 & -2.1e-09 & -1.6e-09 & -4.3e-10 \\
 &  & \texttt{Sgn} & 0.6262 & 0.6282 & 0.6443 & 0.6400 & 0.6022 & 0.5990 & 0.6126 & 0.6177 & 0.5354 & 0.5349 & 0.5462 & 0.5479 \\
\cmidrule{1-15}
 \multirow{12}{*}{\textbf{UMLS}} & \multirow{4}{*}{MT SFT} & \texttt{Cos} & 0.9845 & 0.9821 & 0.9886 & 0.9901 & 0.9819 & 0.9782 & 0.9857 & 0.9875 & 0.9881 & 0.9883 & 0.9920 & 0.9920 \\
 &  & \texttt{Mse} & 2.8e-09 & 2.0e-08 & 1.3e-08 & 1.8e-09 & 2.6e-09 & 1.1e-08 & 7.3e-09 & 1.8e-09 & 1.9e-09 & 1.9e-09 & 1.3e-09 & 1.3e-09 \\
 &  & \texttt{Egy} & 4.4e-06 & 4.4e-07 & 8.0e-07 & 8.3e-06 & -2.8e-06 & 1.8e-06 & 1.8e-07 & 4.8e-06 & -3.2e-10 & 1.1e-09 & -3.8e-09 & -2.8e-09 \\
 &  & \texttt{Sgn} & 0.9457 & 0.9459 & 0.9519 & 0.9531 & 0.9481 & 0.9460 & 0.9516 & 0.9546 & 0.9504 & 0.9502 & 0.9556 & 0.9575 \\
\cmidrule{2-15}
 & \multirow{4}{*}{MT GRPO} & \texttt{Cos} & 0.9898 & 0.9876 & 0.9913 & 0.9922 & 0.9820 & 0.9814 & 0.9870 & 0.9875 & 0.9875 & 0.9877 & 0.9917 & 0.9918 \\
 &  & \texttt{Mse} & 2.0e-09 & 1.4e-08 & 9.7e-09 & 1.5e-09 & 2.1e-09 & 8.9e-09 & 6.2e-09 & 1.5e-09 & 2.0e-09 & 2.0e-09 & 1.3e-09 & 1.3e-09 \\
 &  & \texttt{Egy} & 7.0e-06 & 3.9e-07 & 1.1e-06 & 7.8e-06 & -3.2e-09 & 1.2e-09 & 2.5e-10 & -2.5e-10 & 2.3e-09 & 1.9e-09 & -3.0e-09 & -3.4e-09 \\
 &  & \texttt{Sgn} & 0.9482 & 0.9489 & 0.9544 & 0.9550 & 0.9508 & 0.9489 & 0.9539 & 0.9566 & 0.9507 & 0.9506 & 0.9557 & 0.9575 \\
\cmidrule{2-15}
 & \multirow{4}{*}{MT RM-R1} & \texttt{Cos} & 0.6751 & 0.6835 & 0.7261 & 0.7020 & 0.7014 & 0.6823 & 0.7103 & 0.7238 & 0.4931 & 0.5119 & 0.5722 & 0.5263 \\
 &  & \texttt{Mse} & 5.1e-08 & 3.5e-07 & 3.0e-07 & 4.7e-08 & 3.6e-08 & 1.5e-07 & 1.4e-07 & 3.3e-08 & 8.1e-08 & 7.8e-08 & 6.8e-08 & 7.5e-08 \\
 &  & \texttt{Egy} & 4.4e-06 & 1.7e-06 & 1.7e-06 & 8.5e-06 & 1.7e-10 & -2.1e-09 & 4.1e-10 & -2.6e-09 & 1.9e-09 & -1.8e-09 & 2.3e-09 & 2.4e-09 \\
 &  & \texttt{Sgn} & 0.7332 & 0.7305 & 0.7450 & 0.7487 & 0.7679 & 0.7606 & 0.7660 & 0.7691 & 0.6516 & 0.6505 & 0.6695 & 0.6720 \\
\midrule
\multirow{12}{*}{\textbf{PrimeKG}} & \multirow{4}{*}{MT SFT} & \texttt{Cos} & 0.9872 & 0.9876 & 0.9922 & 0.9918 & 0.9790 & 0.9784 & 0.9856 & 0.9851 & 0.9889 & 0.9893 & 0.9932 & 0.9927 \\
 &  & \texttt{Mse} & 2.0e-09 & 1.3e-08 & 8.5e-09 & 1.3e-09 & 2.5e-09 & 1.0e-08 & 6.9e-09 & 1.8e-09 & 1.8e-09 & 1.7e-09 & 1.1e-09 & 1.2e-09 \\
 &  & \texttt{Egy} & -2.3e-09 & -1.9e-09 & 2.0e-09 & 2.8e-09 & -8.3e-10 & -2.6e-09 & 3.3e-10 & 1.5e-09 & 1.5e-09 & 6.4e-10 & -2.1e-10 & 1.9e-09 \\
 &  & \texttt{Sgn} & 0.9526 & 0.9530 & 0.9588 & 0.9589 & 0.9500 & 0.9478 & 0.9535 & 0.9557 & 0.9536 & 0.9536 & 0.9602 & 0.9607 \\
\cmidrule{2-15}
 & \multirow{4}{*}{MT GRPO} & \texttt{Cos} & 0.9787 & 0.9788 & 0.9869 & 0.9880 & 0.9644 & 0.9625 & 0.9757 & 0.9773 & 0.9920 & 0.9924 & 0.9949 & 0.9944 \\
 &  & \texttt{Mse} & 3.3e-09 & 2.3e-08 & 1.4e-08 & 1.9e-09 & 4.2e-09 & 1.8e-08 & 1.2e-08 & 2.7e-09 & 1.3e-09 & 1.2e-09 & 8.1e-10 & 8.9e-10 \\
 &  & \texttt{Egy} & 1.5e-09 & -4.3e-10 & 3.8e-09 & 1.5e-09 & 1.7e-09 & 9.9e-10 & -8.3e-11 & 2.5e-10 & -9.6e-10 & -1.8e-09 & 1.3e-09 & 5.3e-10 \\
 &  & \texttt{Sgn} & 0.9497 & 0.9496 & 0.9558 & 0.9569 & 0.9449 & 0.9419 & 0.9485 & 0.9519 & 0.9553 & 0.9553 & 0.9620 & 0.9625 \\
\cmidrule{2-15}
 & \multirow{4}{*}{MT RM-R1} & \texttt{Cos} & 0.7520 & 0.7502 & 0.7967 & 0.7738 & 0.7530 & 0.7380 & 0.7681 & 0.7759 & 0.5756 & 0.5928 & 0.6450 & 0.5983 \\
 &  & \texttt{Mse} & 3.9e-08 & 2.7e-07 & 2.2e-07 & 3.5e-08 & 2.9e-08 & 1.2e-07 & 1.1e-07 & 2.7e-08 & 6.7e-08 & 6.5e-08 & 5.6e-08 & 6.4e-08 \\
 &  & \texttt{Egy} & 2.0e-09 & 2.1e-09 & -3.2e-10 & 1.6e-09 & 3.4e-09 & 9.9e-10 & 9.9e-10 & 2.0e-09 & 3.8e-09 & -4.3e-10 & -2.8e-09 & 1.9e-09 \\
 &  & \texttt{Sgn} & 0.7672 & 0.7659 & 0.7793 & 0.7810 & 0.7883 & 0.7822 & 0.7888 & 0.7905 & 0.6740 & 0.6729 & 0.6950 & 0.6973 \\
\bottomrule
\end{tabular}
}
\caption{\small Layer-averaged divergence metrics for attention components (Q, K, V, O) across three models (Qwen2.5-7B, Qwen3-8B, Gemma-7B). We compare the baseline Task SFT against three multi-task paradigms (MT SFT, MT GRPO, MT RM-R1) on two KGs (UMLS, PrimeKG).}
\label{tab:divergence_metrics}
\end{table*}

In this section, we conduct a detailed gradient analysis to compare the fine-tuned model ($\mathbf{g}_B$) against its non fine-tuned baseline ($\mathbf{g}_A$). We utilize four metrics to characterize the optimization trajectory: Cosine Similarity (\textbf{Cos}), Mean Squared Error (\textbf{Mse}), Energy Shift (\textbf{Egy}), and Sign Agreement (\textbf{Sgn}).

\begin{itemize}
    \item \textbf{Cosine Similarity (Cos)} measures the geometric alignment between the fine-tuned and baseline gradient vectors. It is defined as:
    \begin{equation}
        \text{Cos}(\mathbf{g}_A, \mathbf{g}_B) = \frac{\mathbf{g}_A \cdot \mathbf{g}_B}{\|\mathbf{g}_A\| \|\mathbf{g}_B\| + \epsilon},
    \end{equation}
    where $\epsilon$ is a small constant for numerical stability. A value close to $1.0$ indicates that the optimization direction aligns with the baseline manifold.

    \item \textbf{Mean Squared Error (Mse)} quantifies the average magnitude of the difference between updates:
    \begin{equation}
        \text{Mse}(\mathbf{g}_A, \mathbf{g}_B) = \frac{1}{N} \|\mathbf{g}_A - \mathbf{g}_B\|^2,
    \end{equation}
    where $N$ represents the total number of parameters in the layer. Lower MSE values indicate that the model adapts through minimal deviations from the initialization.

    \item \textbf{Energy Shift (Egy)} captures the relative change in gradient magnitude on a logarithmic scale:
    \begin{equation}
    \small
        \text{Egy}(\mathbf{g}_A, \mathbf{g}_B) = \log(\|\mathbf{g}_A\|) - \log(\|\mathbf{g}_B\|).
    \end{equation}
    Negative values indicate that the fine-tuned parameters have a larger magnitude than the baseline, while positive values indicate a smaller magnitude.

    \item \textbf{Sign Agreement (Sgn)} measures the proportion of parameters that retain their original polarity (sign). It is calculated as:
    \begin{equation}
    \small
        \text{Sgn}(\mathbf{g}_A, \mathbf{g}_B) = \frac{1}{N} \sum_{i=1}^{N} \mathbb{I}(\text{sign}(\mathbf{g}_{A,i}) = \text{sign}(\mathbf{g}_{B,i})).
    \end{equation}
    High agreement implies that the update refines existing features rather than overwriting them.
\end{itemize}

Figures \ref{fig:gradient_analysis_part1} and \ref{fig:gradient_analysis_part2}, along with Table \ref{tab:divergence_metrics}, show distinct optimization behaviors across the different training paradigms.
\paragraph{Baseline Instability.}

The Task SFT baseline consistently shows high angular deviation and low Cos ($\text{Cos} \approx 0.11\text{--}0.42$), coupled with a significantly higher Mse compared to all other methods. This suggests a ``brute-force'' approach where the model undergoes aggressive parameter modification to fit the downstream task, likely overwriting pre-trained features essential for general capability.

\paragraph{Surgical Precision in Multi-Task Learning.}
In sharp contrast, our multi-task frameworks, specifically MT SFT and MT GRPO, demonstrate a much more ``surgical'' strategy. These methods maintain near-perfect alignment with the pre-trained model ($\text{Cos} > 0.96$) and achieve high Sgn ($\text{Sgn} > 0.94$). As visualized in the 3D trajectories, they exhibit small Egy, indicating that they integrate KG knowledge through precise, low-rank adjustments rather than wholesale retraining. This stability is evident in Figures \ref{fig:gradient_analysis_part1} and \ref{fig:gradient_analysis_part2}, where the Task SFT often diverges from pre-trained models aggressively, while our methods remain tightly aligned with the pre-trained models.

\paragraph{The Role of Reward Modeling.}
Meanwhile, the MT RM-R1 paradigm shows moderate structural deviation, even without including the explicit KL divergence penalty in our training objective. This intermediate behavior suggests that while reward modeling needs to reshape the feature space, it does so without compromising the model's general knowledge a lot.

\begin{table*}[h!]
\centering
\resizebox{0.95\textwidth}{!}{%
\begin{tabular}{lll cccc cccc cccc }
\toprule
\multirow{2}{*}{\textbf{KG}} & \multirow{2}{*}{\textbf{Method}} & \multirow{2}{*}{\textbf{Threshold}} & \multicolumn{4}{c}{\textbf{Qwen2.5-7B}} & \multicolumn{4}{c}{\textbf{Qwen3-8B}} & \multicolumn{4}{c}{\textbf{Gemma-7B}} \\
\cmidrule(lr){4-7} \cmidrule(lr){8-11} \cmidrule(lr){12-15} 
& & & Q & K & V & O & Q & K & V & O & Q & K & V & O \\
\midrule
\multirow{5}{*}{\textbf{Baseline}} & \multirow{5}{*}{Task SFT} & $10^{-6}$ & 0.00 & 0.00 & 0.00 & 0.00 & 0.00 & 0.00 & 0.00 & 0.00 & 0.00 & 0.00 & 0.00 & 0.00 \\
 &  & $10^{-7}$ & 0.04 & 0.06 & 0.05 & 0.06 & 0.06 & 0.05 & 0.04 & 0.06 & 0.05 & 0.05 & 0.02 & 0.04 \\
 &  & $10^{-8}$ & 0.40 & 0.46 & 0.44 & 0.43 & 0.38 & 0.43 & 0.46 & 0.42 & 0.41 & 0.40 & 0.39 & 0.36 \\
 &  & $10^{-9}$ & 0.40 & 0.46 & 0.44 & 0.43 & 0.38 & 0.43 & 0.46 & 0.42 & 0.41 & 0.40 & 0.39 & 0.36 \\
 &  & $10^{-10}$ & 0.40 & 0.46 & 0.44 & 0.43 & 0.38 & 0.43 & 0.46 & 0.42 & 0.41 & 0.40 & 0.39 & 0.36 \\
\midrule
\multirow{15}{*}{\textbf{UMLS}} & \multirow{5}{*}{MT SFT} & $10^{-6}$ & 0.99 & 0.95 & 0.95 & 0.99 & 0.98 & 0.97 & 0.97 & 1.00 & 0.00 & 0.00 & 0.00 & 0.00 \\
 &  & $10^{-7}$ & 1.00 & 1.00 & 0.99 & 1.00 & 1.00 & 1.00 & 0.99 & 1.00 & 0.03 & 0.05 & 0.05 & 0.04 \\
 &  & $10^{-8}$ & 1.00 & 1.00 & 1.00 & 1.00 & 1.00 & 1.00 & 1.00 & 1.00 & 0.37 & 0.42 & 0.38 & 0.39 \\
 &  & $10^{-9}$ & 1.00 & 1.00 & 1.00 & 1.00 & 1.00 & 1.00 & 1.00 & 1.00 & 0.37 & 0.42 & 0.38 & 0.39 \\
 &  & $10^{-10}$ & 1.00 & 1.00 & 1.00 & 1.00 & 1.00 & 1.00 & 1.00 & 1.00 & 0.37 & 0.42 & 0.38 & 0.39 \\
\cmidrule{2-15}
 & \multirow{5}{*}{MT GRPO} & $10^{-6}$ & 0.99 & 0.94 & 0.92 & 1.00 & 0.00 & 0.00 & 0.00 & 0.00 & 0.00 & 0.00 & 0.00 & 0.00 \\
 &  & $10^{-7}$ & 1.00 & 1.00 & 0.99 & 1.00 & 0.06 & 0.04 & 0.04 & 0.03 & 0.06 & 0.04 & 0.05 & 0.05 \\
 &  & $10^{-8}$ & 1.00 & 1.00 & 1.00 & 1.00 & 0.42 & 0.42 & 0.44 & 0.40 & 0.40 & 0.38 & 0.38 & 0.41 \\
 &  & $10^{-9}$ & 1.00 & 1.00 & 1.00 & 1.00 & 0.42 & 0.42 & 0.44 & 0.40 & 0.40 & 0.38 & 0.38 & 0.41 \\
 &  & $10^{-10}$ & 1.00 & 1.00 & 1.00 & 1.00 & 0.42 & 0.42 & 0.44 & 0.40 & 0.40 & 0.38 & 0.38 & 0.41 \\
\cmidrule{2-15}
 & \multirow{5}{*}{MT RM-R1} & $10^{-6}$ & 0.99 & 0.95 & 0.97 & 0.99 & 0.00 & 0.00 & 0.00 & 0.00 & 0.00 & 0.00 & 0.00 & 0.00 \\
 &  & $10^{-7}$ & 1.00 & 1.00 & 1.00 & 1.00 & 0.06 & 0.05 & 0.04 & 0.07 & 0.08 & 0.04 & 0.04 & 0.05 \\
 &  & $10^{-8}$ & 1.00 & 1.00 & 1.00 & 1.00 & 0.40 & 0.44 & 0.46 & 0.40 & 0.44 & 0.38 & 0.41 & 0.44 \\
 &  & $10^{-9}$ & 1.00 & 1.00 & 1.00 & 1.00 & 0.40 & 0.44 & 0.46 & 0.40 & 0.44 & 0.38 & 0.41 & 0.44 \\
 &  & $10^{-10}$ & 1.00 & 1.00 & 1.00 & 1.00 & 0.40 & 0.44 & 0.46 & 0.40 & 0.44 & 0.38 & 0.41 & 0.44 \\
\midrule
\multirow{15}{*}{\textbf{PrimeKG}} & \multirow{5}{*}{MT SFT} & $10^{-6}$ & 0.00 & 0.00 & 0.00 & 0.00 & 0.00 & 0.00 & 0.00 & 0.00 & 0.00 & 0.00 & 0.00 & 0.00 \\
 &  & $10^{-7}$ & 0.04 & 0.06 & 0.04 & 0.04 & 0.06 & 0.04 & 0.04 & 0.04 & 0.03 & 0.06 & 0.04 & 0.05 \\
 &  & $10^{-8}$ & 0.37 & 0.47 & 0.41 & 0.41 & 0.40 & 0.39 & 0.42 & 0.38 & 0.41 & 0.40 & 0.40 & 0.37 \\
 &  & $10^{-9}$ & 0.37 & 0.47 & 0.41 & 0.41 & 0.40 & 0.39 & 0.42 & 0.38 & 0.41 & 0.40 & 0.40 & 0.37 \\
 &  & $10^{-10}$ & 0.37 & 0.47 & 0.41 & 0.41 & 0.40 & 0.39 & 0.42 & 0.38 & 0.41 & 0.40 & 0.40 & 0.37 \\
\cmidrule{2-15}
 & \multirow{5}{*}{MT GRPO} & $10^{-6}$ & 0.00 & 0.00 & 0.00 & 0.00 & 0.00 & 0.00 & 0.00 & 0.00 & 0.00 & 0.00 & 0.00 & 0.00 \\
 &  & $10^{-7}$ & 0.06 & 0.06 & 0.04 & 0.04 & 0.07 & 0.05 & 0.05 & 0.05 & 0.04 & 0.04 & 0.04 & 0.04 \\
 &  & $10^{-8}$ & 0.42 & 0.42 & 0.41 & 0.37 & 0.41 & 0.44 & 0.41 & 0.42 & 0.41 & 0.37 & 0.39 & 0.38 \\
 &  & $10^{-9}$ & 0.42 & 0.42 & 0.41 & 0.37 & 0.41 & 0.44 & 0.41 & 0.42 & 0.41 & 0.37 & 0.39 & 0.38 \\
 &  & $10^{-10}$ & 0.42 & 0.42 & 0.41 & 0.37 & 0.41 & 0.44 & 0.41 & 0.42 & 0.41 & 0.37 & 0.39 & 0.38 \\
\cmidrule{2-15}
 & \multirow{5}{*}{MT RM-R1} & $10^{-6}$ & 0.00 & 0.00 & 0.00 & 0.00 & 0.00 & 0.00 & 0.00 & 0.00 & 0.00 & 0.00 & 0.00 & 0.00 \\
 &  & $10^{-7}$ & 0.05 & 0.09 & 0.05 & 0.04 & 0.06 & 0.04 & 0.04 & 0.06 & 0.05 & 0.06 & 0.05 & 0.04 \\
 &  & $10^{-8}$ & 0.43 & 0.45 & 0.43 & 0.37 & 0.40 & 0.44 & 0.41 & 0.42 & 0.39 & 0.38 & 0.41 & 0.42 \\
 &  & $10^{-9}$ & 0.43 & 0.45 & 0.43 & 0.37 & 0.40 & 0.44 & 0.41 & 0.42 & 0.39 & 0.38 & 0.41 & 0.42 \\
 &  & $10^{-10}$ & 0.43 & 0.45 & 0.43 & 0.37 & 0.40 & 0.44 & 0.41 & 0.42 & 0.39 & 0.38 & 0.41 & 0.42 \\
\bottomrule
\end{tabular}
}
\caption{Gradient intervention density across various thresholds.}
\label{appendix: gradient_threshold_analysis}
\end{table*}

In general, these findings prove that our KG-guided approach effectively mitigates catastrophic forgetting by updating model parameters in a small task-relevant subspace, allowing the model to adapt well to downstream tasks while preserving its general capabilities.

\end{document}